%% file: main.tex
\pdfoutput=1
\documentclass[11pt]{article}

\usepackage[]{acl}

\usepackage[dvipsnames]{xcolor}
\usepackage{pifont}

\usepackage{times}
\usepackage{latexsym}
\usepackage{graphicx}
\usepackage{adjustbox}
\usepackage{multirow}
\usepackage[normalem]{ulem}
\usepackage{xcolor,colortbl}
\usepackage{soul}
\setuldepth{Berlin}
\usepackage{paralist}
\usepackage{xfrac}
\usepackage[inline]{enumitem}
\usepackage{graphicx}
\usepackage{url}
\usepackage{textcomp}
\usepackage{caption}
\usepackage{subcaption}
\usepackage{float}
\usepackage{placeins}
\usepackage{aliascnt}
\usepackage{mathtools}
\usepackage{amsmath}
\usepackage{hhline}
\usepackage{booktabs}
\usepackage{soul}
\usepackage{bbm}
\usepackage{amssymb}
\usepackage{pifont}
\usepackage{adjustbox}
\usepackage{array}
\usepackage{lipsum}
\usepackage{scrextend}

\usepackage[percent]{overpic}
\usepackage{titlesec}
\usepackage{setspace}
\usepackage{hyperref}
\usepackage{microtype}  
\usepackage{footnote}

\usepackage[T1]{fontenc}

\usepackage[utf8]{inputenc}

\newcommand\blfootnote[1]{%
  \begingroup
  \renewcommand\thefootnote{}\footnote{#1}%
  \addtocounter{footnote}{-1}%
  \endgroup
}

\newcommand{\cmark}{\textcolor{gray}{\scalebox{1.25}{ \ensuremath{\circlearrowright}}}} 

\definecolor{mypink}{rgb}{1.0, 0.552, 0.706}

\definecolor{hlcb}{rgb}{0.941, 0.973, 1.0}
\newcommand{\hlrb}{\rowcolor{hlcb}}

\newcommand{\tmark}{\textcolor{mypink}{\ensuremath{\triangle}}}     

\newcommand{\pmark}{\textcolor{mypink}{\ding{115}}} 

\title{SEA-BED: How Do Embedding Models Represent \\ Southeast Asian Languages?}

\author{
Wuttikorn Ponwitayarat\textsuperscript{1*,$\dagger$}, Peerat Limkonchotiwat\textsuperscript{2*}, Raymond Ng\textsuperscript{2*}, Jann Railey Montalan\textsuperscript{2}, \\
\textbf{Thura Aung\textsuperscript{3}, Jian Gang Ngui\textsuperscript{2}, Yosephine Susanto\textsuperscript{2}, William Chandra Tjhi\textsuperscript{2},} \\
\textbf{Panuthep Tasawong\textsuperscript{1,$\dagger$}, Erik Cambria\textsuperscript{4}, Ekapol Chuangsuwanich\textsuperscript{5}, Sarana Nutanong\textsuperscript{1}} \\
\textsuperscript{1}Vidyasirimedhi Institute of Science and Technology, \textsuperscript{2}AI Singapore, \\
\textsuperscript{3}King Mongkut’s Institute of Technology Ladkrabang, \textsuperscript{4}Nanyang Technological University, \\
\textsuperscript{5}Department of Computer Engineering, Faculty of Engineering, Chulalongkorn University \\
\texttt{\{wuttikorn.p\_s22,panuthep.t\_s20,snutanon\}@vistec.ac.th,} \\
\texttt{\{peerat,raymond,railey,jiangangngui,yosephine,} \\
\texttt{wtjhi\}@aisingapore.org, 66011606@kmitl.ac.th,} \\
\texttt{cambria@ntu.edu.sg, ekapolc@cp.eng.chula.ac.th}
}

\begin{document}
\maketitle
\blfootnote{\textsuperscript{*}Equal contributions} 
\blfootnote{\textsuperscript{$\dagger$}Work was conducted while Wuttikorn Ponwitayarat and Panuthep Tasawong were visiting scholars at AI Singapore}

\blfootnote{Link to SEA-BED: \hyperlink{https://leaderboard.sea-lion.ai/embedding/SEA}{https://leaderboard.sea-lion.ai/embedding/SEA}}

\begin{abstract}

\input{0_Abstract}

\end{abstract}

\input{1_Introduction}

\input{3_SEA-BED_benchmark}

\input{4_Experimental_settings}
\input{5_Results_and_analysis}
\input{2_Related_work}
\input{6_Conclusion}

\bibliography{anthology,custom}
\bibliographystyle{acl_natbib}

\appendix
\input{7_Appendix}

\end{document}

%% file: 0_Abstract.tex
Multilingual text embeddings are often assumed to encode meaning in a perspective-independent semantic space, yielding stable similarity judgments across tasks and languages.
Our results show that this assumption does not hold in practice.
We introduce SEA-BED, a large-scale benchmark covering 10 Southeast Asian (SEA) languages and diverse embedding tasks, designed to systematically examine how embedding performance varies across tasks, languages, and language-task combinations.
Across extensive evaluations, we observe that no single model performs uniformly well across SEA languages; task difficulty differs markedly within languages, and success on one task does not reliably generalize to others. 
Language-task analyses further reveal highly non-uniform performance landscapes, where performance varies across different language-task combinations. 
These findings call for closer attention to performance measurements that provide an expansive view across languages and tasks to uncover inconsistencies in semantic representation. 
Based on these observations, we provide insights for future model development, including data, algorithmic, and architectural considerations.

%% file: 1_Introduction.tex
\section{Introduction}
Text embedding plays a crucial role in NLP by transforming complex linguistic structures into fixed-size vectors that capture semantic locality.
These embeddings are fundamental for various downstream tasks, including semantic textual similarity, retrieval, and re-ranking.
New embedding models have also shifted toward global multilinguality. 
For instance, \citet{wang-etal-2024-improving-text} proposed a multilingual model on Mistral-7B that supports 93 languages, while Jina-embeddings-v3 \cite{sturua2024jinav3embeddings} includes 89 languages.

\begin{table*}[h!]
\renewcommand{\arraystretch}{1}
\vspace{-2mm}
\hspace{-3mm}
\centering
\setlength\doublerulesep{5pt}
\scalebox{0.63}{
\setlength{\tabcolsep}{4pt}
\begin{tabular}{lcccccccc}
\hline
\textbf{Benchmark} & \textbf{\# Languages} & \textbf{\# SEA Languages} & \textbf{\# Datasets} & \textbf{\# Task} & \textbf{\# New datasets} & \textbf{\# SEA datasets} & \textbf{\# Human-Crafted  datasets} \\
& & & & & & & \textbf{(only SEA languages)} \\ \hline \hline
MTEB-French~\cite{DBLP:journals/corr/abs-2405-20468} & 1 & N/A & 18 & 8 & 3 & N/A & N/A \\
C-Pack~\cite{DBLP:conf/sigir/XiaoLZMLN24} & 1 & N/A & 35 & 6 & 35 & N/A & N/A \\
SEB~\cite{DBLP:conf/nips/EnevoldsenKMN24} & 4 & N/A & 24 & 4 & 24 & N/A & N/A \\
MMTEB~\cite{enevoldsen2025mmteb} & 1,090 & 9 & 270 & 10 & 5 & 22 & 21 \\ 
SEA-BED (ours) & 10 & 10 & 169 & 9 & 11 & 169 & 120 (71.01 \%) \\ 
\hline \hline
\end{tabular}
}
\vspace{-2mm}
\caption{The statistics of our benchmark compared to existing text embedding benchmarks.}
\vspace{-5mm}
\label{tab:compared_datasets}
\end{table*}

A common \textbf{ideal} motivating such multilingual development is that a single embedding model should approximate a robust semantic space handling multiple languages simultaneously.
This ideal posits that semantic equivalence should primarily determine geometric similarity, independent of linguistic surface forms, an assumption that underlies the use of multilingual embeddings in cross-lingual retrieval \cite{conneau2018senteval}, semantic search, and transfer learning.
In practice, however, \emph{nothing in modern embedding architectures guarantees such ideal behavior.}
Multilingual text embeddings are learned through co-occurrence statistics in training data. 
It reflects linguistic distributions, translation conventions, cultural norms, and domain-specific usage patterns present in the data. 
Hence, real embedding spaces often diverge across tasks (the ''task consistency problem'') and languages (the ''language consistency problem''), revealing gaps between the ideal of perspective independence and the structure that models actually capture.

Southeast Asia (SEA) presents a uniquely rich environment for exploring the \textbf{real-world limitations} of multilingual text embedding models.
The region spans multiple language families (Austronesian, Tai-Kadai, Sino-Tibetan, Austroasiatic, among others), writing systems (Latin alphabet, abugida, logographic), and typological properties (isolating, analytic, and agglutinative structures).
%
%
Despite the rich epistemic opportunities, \textbf{SEA languages remain underrepresented} in existing embedding benchmarks. 
A practical approach to multilingual benchmarking is translation.
For example, XNLI \cite{conneau2018senteval}, Tatoeba \cite{tatoeba}, and SIB-200 ~\cite{adelani2024sib200simpleinclusivebig} rely on samples translated from English to extend language and task coverage efficiently.
While translation can approximate native data for certain tasks, it may also alter semantic or discourse properties in lower-resource settings. 

An alternative is to use native-authored datasets, which provide stronger linguistic and cultural grounding, motivating MMTEB’s recent focus on provenance nativity~\cite{enevoldsen2025mmteb}.
However, MMTEB includes only 22 SEA datasets across 10 languages, resulting in limited coverage and task diversity.
As a result, current benchmarks provide a narrow view of how embedding models behave in such an epistemically rich testing ground.

\paragraph{Proposed Benchmark.}
When constructing a multilingual benchmark, the first design decision concerns the evaluation scope, the aspects of model behavior we aim to observe. 
In this regard, SEA-BED spans 9 task types across 10 SEA languages, enabling broad cross-task and cross-lingual comparative analysis of multilingual embedding models.
The second decision concerns data sourcing, with the goal of maximizing breadth and depth of coverage while ensuring data quality within the defined scope.
SEA-BED includes 169 datasets, most of which are not covered by existing multilingual sentence embedding benchmarks. 
We combine native-authored datasets with carefully curated translated datasets, enabling systematic comparison between native and transferred provenance. 
We additionally contribute 11 new datasets for Thai and Burmese, enabling deeper examination of semantic similarity, relation understanding, and cross-lingual transfer.
A comparison between SEA-BED and existing benchmarks is given in Table~\ref{tab:compared_datasets}.


\paragraph{Proposed Study.}
With SEA-BED in place, we can systematically examine aspects of multilingual text embeddings that were previously unmeasurable in the SEA region, particularly how model behavior varies across tasks and languages.
\emph{We focus on a single research question: how embedding model behavior varies across tasks and Southeast Asian languages.}
Specifically, we investigate performance patterns under various evaluation conditions. 
To this end, our analysis adopts three complementary comparison views: (i) a language-model view, which analyzes how model behavior varies across languages; (ii) a task-model view, which examines how different embedding models perform across task categories; and (iii) a language-task view, which aggregates model performance over language-task combinations to reveal conditional patterns that emerge at their intersection. 
These three comparative views provide a structured characterization of embedding behavior across languages, tasks, and model dimensions.

\begin{figure*}[ht]
\hspace*{2mm}
\centering
\includegraphics[width=1\textwidth]{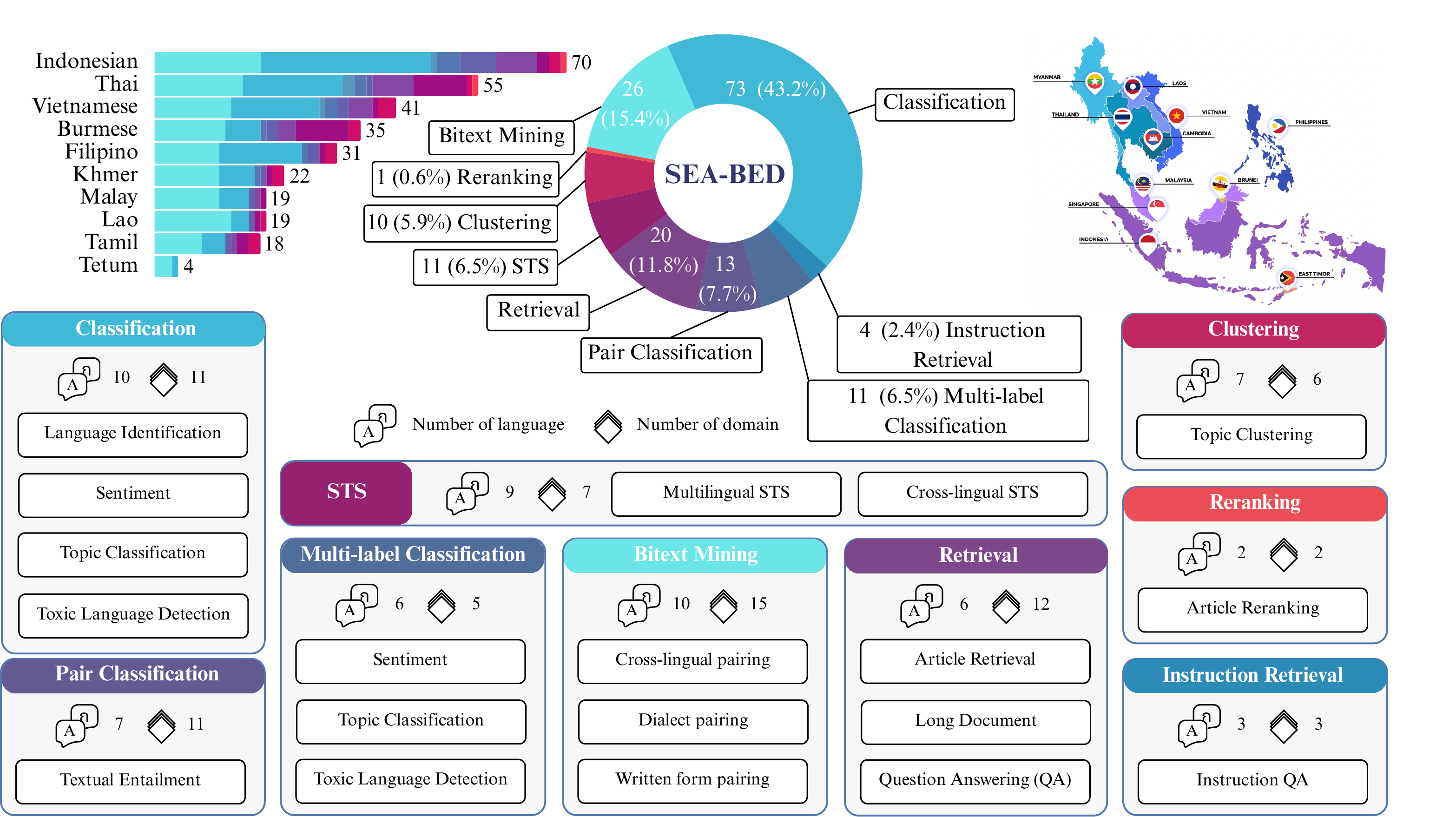}
\vspace{-4mm}
\caption{An overview of SEA-BED, featuring 169 datasets, 9 tasks, and 10 languages.}
\vspace{-6mm}
\label{fig:overview}
\end{figure*}

\paragraph{Key Results.}
Model performance across tasks and SEA languages shows clear drift across languages and task types, with reconfigurations of relative task difficulty rather than uniform scaling, especially in Burmese and Lao. 
Furthermore, there is no single model that performs uniformly well across SEA languages and task types. 
These results indicate that embedding performance is inherently task- and language-dependent.
\textbf{Insights} from our analysis also point to three avenues for performance improvement: enhancing data quality and coverage, refining algorithmic handling of semantic and cross-lingual variation, and adapting architectural designs to better accommodate SEA linguistic diversity.

\paragraph{Our contributions are as follows.}
\begin{compactitem}[\hspace{2mm}•]
    \item \textbf{Resource.} We introduce SEA-BED, a benchmark of 169 datasets across 9 task types and 10 SEA languages, including 11 new Thai and Burmese datasets that expand coverage of previously missing task-language combinations.

    \item \textbf{Experimental Studies.} We evaluate 17 embedding models through a set of studies to analyze performance variation across tasks, languages, and model.

    \item \textbf{Insights for Future Model Development.} Our findings reveal substantial instability in model performance and highlight concrete directions for improving multilingual embedding robustness in SEA settings.
\end{compactitem}


%% file: 3_SEA-BED_benchmark.tex
\section{Proposed Benchmark: SEA-BED}


\subsection{Language-Task Coverage}

We constructed SEA-BED with broad language-task coverage to extend existing evaluation resources for Southeast Asian languages. 
As shown in Figure~\ref{fig:overview}, SEA-BED spans 10 languages and 9 task types, covering linguistic contexts that are substantially underrepresented in benchmarks such as MMTEB. 
%
%
This expanded structure enables the systematic investigation of embedding behavior across tasks and languages.

We adopt the MMTEB task taxonomy~\cite{enevoldsen2025mmteb} to ensure comparability with prior work, and extend their coverage of SEA languages. 
The benchmark spans the following task types. 
%
%
\begin{inparaenum}[\bf (i)]
\item \textbf{Classification}.
Learn a classifier over sentence embeddings to assign labels to individual sentences.
\item \textbf{Multi-label Classification}.
Predict multiple labels for each input text using a classifier trained on embeddings.
%
\item \textbf{Pair Classification}.
Predict a binary relationship between two sentences based on their embedding similarity.
\item \textbf{Semantic Textual Similarity (STS)}.
Measure similarity between sentence pairs using continuous scores derived from distance metrics computed over their embeddings.
%
%
\item \textbf{Clustering}.
The task groups embedded texts into clusters based on semantic similarity, using k-means with the number of unique labels of $k$.
%
%
\item \textbf{Bitext Mining}.
Identify translation pairs across two languages by retrieving the closest match for each sentence in a source set.
\item \textbf{Retrieval}.
Retrieve relevant documents for a given query by computing embedding similarity between the query and candidate texts.
%
\item \textbf{Instruction Retrieval}.
Extend traditional retrieval by incorporating detailed instructions into queries, pairing each query with a corresponding detailed instruction that outlines the criteria for determining document relevance.
\item \textbf{Reranking}.
Reorder a set of candidate documents based on embedding similarity to a query to improve relevance ranking.
%
\end{inparaenum}
Note that summarization is omitted due to data availability constraints. 
%

Table~\ref{tab:task-coverage} summarizes the task subtypes and their language coverage. 
SEA-BED contains 19 subtypes across 9 task types and substantially increases the number of language-subtype pairs relative to MMTEB (from 10 to 19). 
Notably, several subtypes \emph{Language Identification, Toxic Language Detection, Dialect Pairing, Instruction QA, and Article Reranking} are introduced for the first time in SEA language evaluations.

\begin{table}[htbp]
  \centering
  \hspace{-2mm}
  \scalebox{0.78}{
  \setlength{\tabcolsep}{2pt}
  \renewcommand{\arraystretch}{1.15}
  \small
  \renewcommand{\quad}{\hspace{0.3em}}

  \begin{tabular}{lcccccccccc}
    \toprule
    Task Type and Subtype & ind & tha & vie & mya & fil & khm & zsm & lao & tam & tet \\
    \midrule
    \hlrb\textbf{Classification} & & & & & & & & & & \\
    \hlrb\quad Language Identification
      & \pmark & \pmark & \pmark & \pmark & \pmark & \pmark & \pmark & \pmark &        & \pmark \\
    \hlrb\quad Sentiment
      & \cmark & \cmark & \cmark & \pmark & \cmark & \pmark &        &        & \tmark &        \\
    \hlrb\quad Topic Classification
      & \cmark & \cmark & \cmark & \pmark & \cmark & \cmark & \cmark & \tmark & \cmark &        \\
    \hlrb\quad Toxic Language Detection
      & \pmark & \pmark & \pmark &        & \pmark &        &        &        &        &        \\

    \hlrb\textbf{Multi-label Classification}  & & & & & & & & & & \\
    \hlrb\quad Sentiment
      & \pmark &        & \pmark &        &        &        &        &        &        &        \\
    \hlrb\quad Topic Classification
      &        & \pmark &        & \pmark & \pmark & \pmark &        &        &        &        \\
    \hlrb\quad Toxic Language Detection
      & \pmark &        &        &        &        &        &        &        &        &        \\

    \hlrb\textbf{Pair Classification}  & & & & & & & & & & \\
    \hlrb\quad Textual Entailment
      & \cmark & \cmark & \cmark & \pmark & \pmark & \pmark & \pmark & \pmark & \pmark &        \\
    \addlinespace[0.3em]

    \textbf{STS}  & & & & & & & & & & \\
    \quad Multilingual STS
      & \cmark & \pmark & \pmark & \pmark & \pmark & \pmark & \pmark & \pmark & \pmark &        \\
    \quad Cross-lingual STS
      &        & \pmark &        & \pmark &        &        &        &        & \cmark &        \\
    \addlinespace[0.3em]

    \textbf{Clustering}  & & & & & & & & & & \\
    \quad Topic Clustering
      & \cmark & \cmark & \cmark & \cmark & \cmark & \cmark & \cmark & \cmark & \tmark &        \\
    \addlinespace[0.3em]

    \textbf{Bitext Mining}  & & & & & & & & & & \\
    \quad Cross-lingual pairing
      & \cmark & \cmark & \cmark & \cmark & \cmark & \cmark & \tmark & \cmark & \cmark & \pmark \\
    \quad Dialect pairing
      & \pmark & \pmark & \pmark & \pmark & \pmark & \pmark & \pmark & \pmark &        &        \\
    \quad Written-forms pairing
      & \pmark & \pmark & \pmark & \pmark &        &        &        &        &        &        \\
    \addlinespace[0.3em]

    \hlrb\textbf{Retrieval}  & & & & & & & & & & \\
    \hlrb\quad Article Retrieval
      & \cmark & \cmark & \pmark & \pmark &        &        &        &        &        &        \\
    \hlrb\quad Long Document Retrieval
      &        & \pmark &        &        &        &        &        &        &        &        \\
    \hlrb\quad Question Answering
      & \pmark & \pmark & \cmark &        &        &        & \pmark &        & \pmark &        \\

    \hlrb\textbf{Instruction Retrieval}  & & & & & & & & & & \\
    \hlrb\quad Instruction QA
      & \pmark & \pmark & \pmark &        &        &        &        &        &        &        \\

    \hlrb\textbf{Reranking}  & & & & & & & & & &  \\
    \hlrb\quad Article Reranking
      & \pmark & \pmark &        &        &        &        &        &        &        &        \\
    \bottomrule
  \end{tabular}
  }
  \vspace{-2mm}
  \caption{Task coverage of SEA-BED compared to MMTEB. Cells use {\cmark} (present in MMTEB and directly reused), {\tmark} (present in MMTEB and extended in SEA-BED), and {\pmark} (entirely new). Similar task types are grouped by \colorbox{hlcb}{highlight}.}
  \vspace{-5mm}
  \label{tab:task-coverage}
\end{table}


\subsection{Evaluation Data Sourcing}

\paragraph{Data Sourcing Overview.}
We now turn our attention to how datasets were sourced and organized to enable systematic evaluation within the scope described in the previous subsection.
%
%
SEA-BED consists of 169 datasets spanning 9 task types and 10 languages, as previously discussed in Table~\ref{tab:compared_datasets}.
While MMTEB contains 270 datasets, only 22 involve SEA languages, which reflects a limited regional representation. 
In contrast, 147 of our datasets (86\%) are not included in MMTEB, highlighting substantial gaps in existing benchmark coverage, making SEA-BED substantially more representative of SEA linguistic diversity.
Of these, 120 were authored by native speakers in their respective languages, while the rest are sourced from translated datasets. 
This combination enables the study of provenance effects across task types and languages.
We also introduce 11 new datasets for Thai and Burmese to expand coverage of previously underrepresented task-language combinations. Benchmark efficiency considerations (e.g., caching and downsampling) are described in Appendix~\ref{appendix:benchmark_efficiency}.

\paragraph{Domain Coverage.}
Domain diversity is essential for realistic and discriminative evaluation, as semantic representations vary substantially across domains and influence model robustness in real-world applications.
MMTEB provides coverage of widely studied domains in SEA languages (e.g., news, non-fiction, and encyclopedias).
SEA-BED complements and extends this foundation by spanning 17 domains across 10 SEA languages, including a broader range of formal, informal, and application-oriented texts.
Much of this coverage is supported by new or newly sourced datasets, with some domains (academic, blogs, medical, and subtitles) newly introduced.
Table~\ref{tab:domain_comparison} presents the domain coverage of the SEA-BED benchmark compared to MMTEB across languages, indicating reused, extended, and newly introduced domains.
The full domain descriptions are provided in Appendix~\ref{appendix:domain_details}.
\begin{table}[h!]
\hspace{-3mm}
\centering
\scalebox{0.65}{
\setlength{\tabcolsep}{5pt}
\renewcommand{\arraystretch}{1.2}
\begin{tabular}{lcccccccccc}
    \toprule
    \textbf{Domain} & \textbf{ind} & \textbf{tha} & \textbf{vie} & \textbf{mya} & \textbf{fil} & \textbf{khm} & \textbf{zsm} & \textbf{lao} & \textbf{tam} & \textbf{tet} \\ 
    \midrule
    Academic &  &  & \pmark &  &  &  &  &  &  &  \\
    Blog & \pmark & \pmark & \pmark & \pmark & \pmark &  &  &  &  &  \\
    Constructed & \pmark & \pmark & \pmark & \pmark & \pmark &  & \pmark &  &  &  \\
    Encyclopedia & \tmark & \tmark & \tmark & \tmark & \tmark & \tmark & \pmark & \tmark & \cmark &  \\
    Fiction & \pmark & \cmark & \tmark & \pmark & \cmark & \pmark & \pmark & \pmark & \cmark &  \\
    Government & \pmark & \tmark & \tmark & \pmark & \pmark & \pmark & \pmark & \pmark & \cmark & \pmark \\
    Legal & \pmark & \pmark & \pmark & \pmark &  & \pmark & \pmark & \pmark & \cmark &  \\
    Medical &  & \pmark & \pmark &  &  &  &  &  &  &  \\
    News & \tmark & \tmark & \tmark & \tmark & \tmark & \tmark & \tmark & \tmark & \tmark &  \\
    Non-fiction & \tmark & \tmark & \tmark & \tmark & \tmark & \tmark & \pmark & \tmark & \cmark &  \\
    Religious & \tmark & \cmark & \cmark & \cmark & \cmark & \pmark & \pmark & \pmark & \cmark & \pmark \\
    Reviews & \tmark & \tmark & \tmark &  & \pmark & \pmark & \pmark &  &  &  \\
    Social & \pmark & \pmark & \pmark & \pmark & \tmark &  &  &  & \cmark &  \\
    Spoken & \tmark & \tmark & \tmark & \pmark & \tmark & \tmark & \tmark & \tmark & \tmark & \pmark \\
    Subtitles & \pmark &  &  &  &  &  &  &  &  &  \\
    Web & \tmark & \pmark & \pmark & \pmark &  &  & \pmark &  & \cmark & \pmark \\
    Written & \tmark & \tmark & \tmark & \tmark & \tmark & \tmark & \pmark & \tmark & \tmark & \pmark \\
    \bottomrule
\end{tabular}
}
\vspace{-1mm}
\caption{Domain coverage of SEA-BED benchmark compared to MMTEB. Cells use {\cmark} to denote tasks present in MMTEB and directly reused, {\tmark} for tasks present in MMTEB but extended in SEA-BED, and {\pmark} for entirely new domains.}
\vspace{-3mm}
\label{tab:domain_comparison}
\end{table}

\paragraph{Data Quality Assurance.} 
%
%
%
Data quality assurance is achieved through systematic review processes conducted by native speakers of SEA languages, all of whom are also proficient in English. 
These annotators verified and validated the data for grammatical correctness, native written style, appropriate language usage (excluding code-switching), and the accuracy of the gold standard annotations.
This human verification process results in approximately 8\% of the datasets being removed, reducing the number from 182 to 169 datasets.
See Appendix~\ref{appendix:annotator_demographics} for the annotator guidelines and details.

%
%

\paragraph{New Datasets.}
In addition to curated datasets, we construct new Thai and Burmese datasets for semantic textual similarity, natural language inference, and multi-label classification. 
These new resources address task-level gaps in SEA languages and expand evaluation coverage in previously underrepresented settings.
Given the importance of these tasks for downstream retrieval and re-ranking performance~\cite{simcse,diffcse}, we release 4 Thai datasets comprising 3,147 samples and 7 Burmese datasets with 13,177 samples to support systematic evaluation.

As shown in Table~\ref{tab:new_dataset_statistic}, we construct these datasets by human verification of Google NMT translated data from established English benchmarks for semantic textual similarity and natural language inference, including STSBenchmark~\cite{cer-etal-2017-semeval}, STS-2017~\cite{cer-etal-2017-semeval}, STS-2022~\cite{chen-etal-2022-semeval}, STS-2024~\cite{ousidhoum-etal-2024-semeval}, BIOSSES~\cite{souganciouglu2017biosses}, and XNLI~\cite{conneau-etal-2018-xnli}, which serve as the original texts for dataset construction.
We also translate the Thai multi-label dataset Prachathai67k~\cite{prachathai67k} into Burmese,\footnote{Thai is a more culturally and politically compatible source for Burmese translation than English.} providing a stronger starting point for creating Burmese resources.
Translations were performed by native Thai and Burmese speakers (see Appendix~\ref{appendix:annotator_demographics} for annotator demographics and guidelines) following two instructions: (i) produce natural, conversational language, and (ii) make sentence subjects gender-neutral, given that both languages encode gender morphologically.
%

%
%

\begin{table}[h!]
\renewcommand{\arraystretch}{1}
\hspace*{-3mm}
\centering
\setlength\doublerulesep{5pt}
\scalebox{0.8}{
\setlength{\tabcolsep}{8pt}
\begin{tabular}{lcc}
        \toprule
        \textbf{Dataset} & \textbf{mya} & \textbf{tha} \\
        \midrule
        Biosses & 100 & 100 \\
        STS17 & 250 & 250 \\
        STS22 & 197 & 197 \\
        STS24 & 2,600 & 2,600 \\
        STSBenchmark & 2,880 & - \\
        XNLI & 5,000 & - \\
        Prachathai67k & 2,150 & - \\
        \midrule
        Total number of samples & 13,177 & 3,147 \\
        \midrule
    \end{tabular}
}
\vspace{-1mm}
\caption{Statistics of the new evaluation datasets included in SEA-BED.}
\vspace{-3mm}
\label{tab:new_dataset_statistic}
\end{table}

\begin{table*}[h!]
\hspace*{-3mm}
\centering
\setlength\doublerulesep{4pt}
\scalebox{0.75}{
\setlength{\tabcolsep}{5pt}
    \renewcommand{\arraystretch}{1.2}
    \begin{tabular}{lccccccccccc}
        \toprule
        \textbf{Model} & \textbf{ind} & \textbf{tha} & \textbf{vie} & \textbf{mya} & \textbf{fil} & \textbf{khm} & \textbf{zsm} & \textbf{lao} & \textbf{tam} & \textbf{tet} & \textbf{Avg.} \\
        \bottomrule
        \midrule
        \textit{Number of datasets (→)} & (70) & (55) & (41) & (35) & (31) & (22) & (19) & (19) & (18) & (4) & (314) \\
        \midrule

        multilingual-e5-large-instruct (560M)
        & 79.50 & \underline{81.11} & 78.00 & \textbf{78.37} & \underline{79.19} & \textbf{78.13} & \textbf{84.60} & \textbf{83.94} & 77.09 & \underline{69.40} & \textbf{78.93}$_{\pm3.98}$ \\

        Qwen3-Embedding-8B (8B)
        & 79.73 & \textbf{81.49} & \textbf{78.99} & \underline{74.91} & 78.05 & 75.46 & 82.39 & 78.20 & 75.95 & 67.44 & \underline{77.26}$_{\pm4.02}$ \\

        bge-m3 (568M)
        & 78.09 & 77.59 & 75.91 & 73.12 & 75.78 & \underline{76.23} & 82.54 & \underline{82.26} & 77.51 & 65.53 & 76.46$_{\pm4.55}$ \\

        multilingual-e5-large (560M)
        & 78.59 & 79.89 & \underline{78.93} & 70.28 & 77.98 & 72.11 & 80.10 & 79.91 & \underline{77.83} & 63.55 & 75.92$_{\pm5.22}$ \\
        
        bge-multilingual-gemma2 (9B)
        & \underline{79.93} & 80.58 & 78.76 & 70.01 & \textbf{79.61} & 74.39 & \underline{83.38} & 65.82 & \textbf{80.96} & 65.05 & 75.85$_{\pm6.31}$ \\

        LaBSE (471M)
        & 73.98 & 70.20 & 72.60 & 73.63 & 76.99 & 74.06 & 82.87 & 79.84 & 76.59 & 69.11 & 74.99$_{\pm3.99}$ \\

        multilingual-mpnet-base (278M)
        & 74.60 & 73.91 & 72.66 & 61.19 & 52.02 & 64.44 & 75.48 & 65.63 & 63.31 & 50.78 & 65.40$_{\pm8.53}$ \\

        e5-mistral-7b-instruct (7B)
        & 79.23 & 74.77 & 75.37 & 48.85 & 78.10 & 56.49 & 78.82 & 27.99 & 66.73 & 66.73 & 65.32$_{\pm15.74}$ \\

        GritLM-7B (7B)
        & \textbf{80.47} & 72.84 & 77.37 & 45.05 & 77.49 & 52.58 & 78.41 & 30.07 & 60.42 & \textbf{69.67} & 64.44$_{\pm16.13}$ \\ 

        Qwen3-Embedding-0.6B (595M)
        & 75.60 & 75.85 & 75.13 & 49.08 & 63.11 & 44.10 & 69.51 & 29.78 & 61.12 & 63.38 & 60.67$_{\pm14.55}$ \\

        multilingual-MiniLM-L12 (118M)
        & 71.48 & 70.42 & 69.90 & 54.48 & 47.28 & 39.92 & 69.58 & 45.34 & 27.88 & 47.69 & 54.40$_{\pm14.53}$ \\

        Gemma-SEA-LION-v3-9B-IT (9B)
        & 49.86 & 41.67 & 51.90 & 30.80 & 54.14 & 39.53 & 49.24 & 22.01 & 29.20 & 25.06 & 39.34$_{\pm11.27}$ \\

        Sailor2-8B-Chat (8B)
        & 49.54 & 35.98 & 42.94 & 30.14 & 46.16 & 28.57 & 30.75 & 18.31 & 28.57 & 25.76 & 33.67$_{\pm9.33}$ \\

        \midrule
        \emph{Proprietary models} \\ \midrule

        embed-multilingual-v3.0
        & \textbf{79.72} & \textbf{80.99} & \textbf{78.93} & \textbf{76.13} & \textbf{78.99} & \textbf{77.01} & \textbf{82.42} & \textbf{83.34} & \textbf{78.87} & \textbf{66.76} & \textbf{78.32}$_{\pm4.39}$ \\

        jina-embeddings-v3
        & 77.35 & \underline{78.64} & \underline{76.10} & \underline{75.10} & \underline{74.25} & \underline{74.73} & \underline{77.91} & \underline{77.91} & \underline{76.14} & \underline{65.11} & \underline{75.32}$_{\pm3.68}$ \\

        voyage-3
        & 75.56 & 69.78 & 73.68 & 48.19 & 71.43 & 35.02 & 69.13 & 24.27 & 67.28 & 61.48 & 59.58$_{\pm16.83}$ \\ 

        text-embedding-3-small
        & \underline{78.34} & 55.24 & 70.06 & 32.79 & 68.08 & 30.15 & 69.78 & 23.97 & 35.38 & 65.09 & 52.89$_{\pm19.18}$ \\

        \bottomrule
    \end{tabular}
    }
    \vspace{-2mm}
    \caption{Language-model performance view, where each cell reports scores averaged over all evaluated task types.}
    \vspace{-5mm}
    \label{tab:sea-mteb_language_results}
\end{table*}

%% file: 4_Experimental_settings.tex
\section{Experimental Settings}

\noindent
\textbf{Models.}
We evaluate 13 open-source and 4 proprietary multilingual text embedding models on SEA-BED, spanning both encoder-based and decoder-based architectures. 
%
%
%
Our model selection aims to encompass representative design choices in contemporary multilingual embedding models, rather than providing an exhaustive comparison or establishing a single best-performing approach. 

All models are treated as black-box embedding functions, and our analysis focuses on observed performance variation across tasks and languages. Detailed model descriptions, training backgrounds, and per-model results are provided in Appendix~\ref{appendix:models}.

\noindent
\textbf{Evaluation Setup.}
We employ F1 for Bitext Mining, Classification, and Multi-label Classification.
For Pair Classification, we use average precision (AP) as the main metric.
For Clustering, we use the V-measure metric.
For Retrieval, we use various metrics (i.e., nDCG@k, MRR@k, MAP@k, precision@k, and recall@k), with nDCG@10 as the primary metric.
For Reranking, we use Mean Average Precision (MAP). 
In addition, we use nDCG@5 as the main metric for Instruction Retrieval following~\citet{weller2024followir}.
We employ the averaging strategy similar to previous works~\cite{Muennighoff2023mteb,enevoldsen2025mmteb}, where all tasks are averaged equally with the standard deviation (SD) score.
We acknowledge that the metrics for each task are different (e.g., F1 for classification and nDCG@10 for retrieval).
Thus, we analyze both individual and average results, rather than focusing solely on the average score.
%
%
All experiments were run on eight H100 (80 GB). 

%% file: 5_Results_and_analysis.tex
\section{Experimental Results}
\label{sec:exp}

%
%


\begin{table*}[h]
\vspace{-1mm}
\hspace*{-3.5mm}
\centering
\setlength\doublerulesep{4pt}
\scalebox{0.68}{
\setlength{\tabcolsep}{5pt}
\renewcommand{\arraystretch}{1.2}
\begin{tabular}{lc|cccccccccc}
\toprule
\textbf{Model} & \textbf{Dim.} &
\textbf{Clf} & \textbf{M. Clf} & \textbf{Pr. Clf} & \textbf{STS} &
\textbf{Clust} & \textbf{Btxt} & \textbf{Rtrvl} & \textbf{In. Rtrvl} & \textbf{Rrnk} & \textbf{Avg.} \\
\bottomrule
\midrule
\textit{Number of datasets (→)} &  &
(73) & (11) & (13) & (11) & (10) & (26) & (20) & (4) & (1) & (169) \\
\midrule

multilingual-e5-large-instruct (560M) & 1024 &
77.70 & 87.84 & 66.58 & \textbf{75.59} &
\textbf{58.09} & \textbf{87.86} & 77.16 & 69.10 & 77.24 & \textbf{75.24}$_{\pm9.06}$ \\

Qwen3-Embedding-8B (8B) & 4096 &
\textbf{78.60} & \underline{90.57} & 63.10 & \underline{75.31} &
\underline{52.93} & 84.78 & \textbf{81.99} & \underline{70.81} & \underline{78.51} & \underline{75.18}$_{\pm10.84}$ \\

bge-multilingual-gemma2 (9B) & 3584 &
78.13 & \textbf{90.89} & \textbf{73.87} & 72.53 &
49.14 & 82.02 & \underline{80.55} & \textbf{71.52} & 69.04 & 74.19$_{\pm10.85}$ \\

multilingual-e5-large (560M) & 1024 &
\underline{78.24} & 88.94 & 65.79 & 69.61 &
47.83 & 84.51 & 78.25 & 66.06 & \textbf{79.00} & 73.14$_{\pm11.66}$ \\

bge-m3 (568M) & 4096 &
75.98 & 89.89 & 68.73 & 73.27 &
42.23 & 86.18 & 73.56 & 58.51 & 75.98 & 71.59$_{\pm13.48}$ \\

GritLM-7B (7B) & 4096 &
77.47 & 88.76 & 63.86 & 64.69 &
46.29 & 63.63 & 65.97 & 67.60 & 73.37 & 67.96$_{\pm10.92}$ \\

e5-mistral-7b-instruct (7B) & 4096 &
76.65 & 88.32 & 63.81 & 63.50 &
49.48 & 65.30 & 72.93 & 54.46 & 75.33 & 67.75$_{\pm11.24}$ \\

Qwen3-Embedding-0.6B (595M) & 1024 &
74.47 & 88.19 & 60.36 & 65.74 &
43.94 & 56.53 & 76.24 & 65.80 & 75.03 & 67.37$_{\pm11.58}$ \\

multilingual-mpnet-base (278M) & 768 &
73.79 & 87.28 & \underline{70.79} & 70.15 &
41.12 & 68.12 & 58.28 & 52.44 & 64.01 & 65.11$_{\pm12.55}$ \\

LaBSE (471M) & 768 &
75.19 & 86.65 & 62.32 & 68.32 &
41.39 & \underline{86.84} & 53.72 & 39.73 & 61.23 & 63.93$_{\pm16.31}$ \\

multilingual-MiniLM-L12 (118M) & 768 &
70.50 & 84.88 & 65.70 & 64.59 &
31.50 & 53.23 & 52.47 & 48.66 & 62.27 & 59.31$_{\pm14.25}$ \\

Gemma-SEA-LION-v3-9B-IT (9B) & 3584 &
75.87 & 89.94 & 57.77 & 38.85 &
39.94 & 15.31 & 22.03 & 11.02 & 65.49 & 46.25$_{\pm26.18}$ \\

Sailor2-8B-Chat (8B) & 3584 &
76.43 & 90.21 & 56.71 & 37.25 &
38.51 & 4.31 & 10.09 & 3.29 & 47.05 & 40.43$_{\pm29.25}$ \\

\midrule
\emph{Proprietary models} \\ \midrule

embed-multilingual-v3.0 & 1024 &
\textbf{78.52} & \textbf{89.98} & \textbf{66.11} & \underline{73.11} &
\underline{48.99} & \textbf{88.32} & \textbf{78.17} & \underline{65.59} & \textbf{77.77} & \textbf{74.06}$_{\pm11.89}$ \\

jina-embeddings-v3 & 1024 &
\underline{77.40} & \underline{88.97} & \underline{63.61} & \textbf{73.17} &
\textbf{50.90} & \underline{81.86} & \underline{76.28} & \textbf{69.11} & 72.49 & \underline{72.64}$_{\pm10.30}$ \\

voyage-3 & 1024 &
75.72 & 88.70 & 60.23 & 61.97 &
45.15 & 55.62 & 62.91 & 61.77 & \underline{74.62} & 65.19$_{\pm12.01}$ \\

text-embedding-3-small & 1536 &
72.88 & 88.19 & 60.16 & 52.31 &
39.34 & 43.12 & 65.18 & 52.87 & 71.25 & 60.59$_{\pm14.65}$ \\

\bottomrule
\end{tabular}
}
\vspace{-3mm}
\caption{Task-model performance view, where each cell reports scores averaged over all evaluated languages.}
\vspace{-3mm}
\label{tab:sea-mteb_task_results}
\end{table*}

\subsection{Language-Model Comparisons} \label{subsec:lang_model}
%

%
This study presents a language-wise analysis for behavior across language and model to pinpoint which SEA languages and scripts see the largest performance drops.
Note that the number of datasets can be duplicated in multilingual scenarios, resulting in a higher number of datasets than the task-model comparison (Table~\ref{tab:sea-mteb_task_results}).

\noindent
\textbf{Results}.
As shown in Table~\ref{tab:sea-mteb_language_results}, we observe performance variation across languages that, when compared to the task-level results in Table~\ref{tab:sea-mteb_task_results}, suggests that strong overall performance does not necessarily imply consistent multilingual coverage.
For example, while multilingual-e5-large-instruct achieves the highest average score overall (78.93), its performance varies across languages, ranging from 69.40 points in Tetum to 84.60 points in Malay.
We also observe the same trend in proprietary models.
Moreover, we found that \emph{some models do not fully support SEA languages}, e.g., GritLM-7B does not support Burmese, Khmer, and Lao, while bge-multilingual-gemma2 does not support Lao. 
This is because the training datasets for these languages are smaller and of lower quality. 
Although previous works demonstrate the use of LLMs to generate more datasets~\cite{qwen3embedding,muennighoff2024generative}, applying this methodology to low-resource languages is underexplored and might not be effective.
Thus, although these models performed well overall in this experimental study, their lack of support for some SEA languages renders them less suitable for real-world applications involving SEA languages.

\noindent
\textbf{Discussion}. 
Experimental results demonstrate the ``language consistency'' problem, where models exhibit inconsistent performance across different languages. 
Although multilingual-e5-large-instruct might perform best overall, we found that no model can perform best for all languages.
We observe that while multilingual-e5-large-instruct performs well on Burmese, Khmer, Malay, and Lao, Qwen3-Embedding-8B performs well on Thai and Vietnamese, bge-multilingual-gemma2 performs well on Filipino and Tamil, and GritLM-7B performs well on Indonesian and Tetum.
This emphasizes the inconsistency of model performance across languages, thus rendering the overall evaluation results inconclusive for all models.
Making the embedding model consistent to support all languages equally is a challenge and remains an open question in the field of text understanding.
Note that we also experimented with tokenizer and language similarity to understand the underrepresented languages in Appendix~\ref{appedix:tokenizer_analysis} and Appendix~\ref{appedix:lang_correlation}, respectively.

\subsection{Task-Model Comparisons} \label{subsec:task_model}

To understand the performance of each task, we ask which tasks remain particularly challenging for state-of-the-art models across SEA languages. 
%
%

\noindent
\textbf{Results}. As shown in Table~\ref{tab:sea-mteb_task_results}, the experiment results demonstrate that multilingual-e5-large-instruct performs the best on our benchmark, achieving 75.24 points on the average score.
The performance of the second-best model (Qwen3-Embedding-8B) is lower than that of multilingual-e5-large-instruct by only 0.06 points on average, with a 70-fold difference in model parameters (560M vs. 8B parameters).
Moreover, we found that, although Gemma-SEA-LION-v3 and Sailor2 were specifically trained for SEA languages, the models did not perform well on our text embedding benchmark due to their design for generation, rather than embedding purposes.
For the proprietary models, in contrast to previous works~\cite{Muennighoff2023mteb}, which found that proprietary models outperformed open-source models, we found that all proprietary models perform lower than multilingual-e5-large-instruct and Qwen3-Embedding-8B.
This suggests that all proprietary models may be primarily trained in English and not optimized for SEA languages.

\noindent
\textbf{Discussion}. 
We found that task performance consistency is the main challenge for current text embedding models.
In particular, a robust model should perform well on all tasks.
As shown in Table~\ref{tab:sea-mteb_task_results}, we found that there is no dominant model that achieves the highest score on all tasks. 
Notably, model performance varies considerably depending on the task.
%
%
\emph{This emphasizes that the task consistency problem in our benchmark is still challenging for embedding models}.
In short, when using multilingual text embedding in SEA languages, it is essential to select the model based on the specific task at hand, as there is no all-purpose model that suits every solution.

\subsection{Language-Task Comparisons}
\label{subsec:task_lang}


Let us now turn our attention to language-task comparisons, averaging performance across all models.

\noindent
\textbf{Results.}
Table~\ref{tab:sea-mteb_language_task_results} presents our third comparative view: language-task. 
Substantial variation is observed across both dimensions, with no task exhibiting uniformly strong performance across all languages.
%
%
For relatively well-studied tasks such as classification and bitext mining, performance is generally strong for higher-resource languages (ind and tha), while noticeable degradation persists for more resource-constrained languages, e.g., lao, mya, and tet.
%
%
In contrast, clustering remains challenging across most languages, exhibiting lower and more variable performance, indicating that some task categories pose intrinsic difficulties that are not alleviated by language coverage alone.
%
%
Bitext mining exhibits mixed behavior, with strong performance in some languages but substantial variation across others, highlighting how task difficulty interacts with language-specific factors.

\noindent
\textbf{Discussion.}
The results reveal a landscape of conditional behaviors across tasks and languages, underscoring the importance of disentangling evaluation dimensions for meaningful assessment.

\begin{table}[ht]
\hspace*{-3.5mm}
\centering
\setlength\doublerulesep{4pt}
\scalebox{0.58}{
\setlength{\tabcolsep}{5pt}
\renewcommand{\arraystretch}{1.2}
\begin{tabular}{l|ccccccccc}
\toprule
\textbf{Lang.} &
\textbf{Clf} & \textbf{M. Clf} & \textbf{Pr. Clf} & \textbf{STS} &
\textbf{Clust} & \textbf{Btxt} & \textbf{Rtrvl} & \textbf{In. Rtrvl} & \textbf{Rrnk} \\
\bottomrule
\midrule
ind & 77.77 & 92.49 & 64.51 & 61.04 & 43.55 & 75.65 & 72.71 & 34.82 & 69.28 \\
tha & 73.73 & 76.60 & 70.31 & 68.73 & 38.49 & 70.99 & 67.71 & 70.43 & 71.86 \\
vie & 78.93 & 96.49 & 63.67 & 77.15 & 38.56 & 76.36 & 56.80 & 42.75 & - \\
mya & 78.78 & 69.71 & 67.10 & 61.95 & 28.09 & 48.83 & 55.44 & - & - \\
fil & 72.78 & 90.06 & 50.63 & 68.41 & 52.69 & 69.56 & - & - & - \\
khm & 68.56 & 100.00 & 66.41 & 63.74 & 23.83 & 54.21 & - & - & - \\
zsm & 72.94 & - & 71.50 & 75.39 & - & 75.98 & 46.27 & - & - \\
lao & 70.55 & - & 64.36 & 59.11 & 18.23 & 51.79 & - & - & - \\
tam & 84.35 & - & 68.14 & 52.49 & 38.75 & 61.23 & 46.27 & - & - \\
tet & 99.81 & - & - & - & - & 45.75 & - & - & - \\
\bottomrule
\end{tabular}
}
\vspace{-3mm}
\caption{Language-task performance view, where each cell reports scores averaged over all evaluated models.}
\label{tab:sea-mteb_language_task_results}
\vspace{-3mm}
\end{table}

\section{Insights for Future Model Development} \label{sec:insights}
From the main results and ablation studies, embedding models for SEA languages can be enhanced in three aspects: (i) datasets, (ii) training algorithm, and (iii) architecture.
%

\noindent
\textbf{Dataset.} 
A common technique to improve downstream tasks is to introduce data aligned with the domain of the target task.
However, in resource-constrained settings, model developers often have to resort to machine-assisted dataset generation.
As shown in Appendix~\ref{appendix:translation_vs_human}, we examine the possibility of using MT on low-resource languages and found that we can use machine translation to translate from English to SEA languages with a marginal difference between human and machine translations. 
There are various English datasets that are not yet available in SEA languages, i.e., MSMACO~\cite{DBLP:journals/corr/NguyenRSGTMD16} and NQ~\cite{47761}, and some datasets are only available in a subset of SEA languages, e.g., only Thai or Indonesian, like Mr.TyDi~\cite{tydiqa} and MIRACL~\cite{10.1162/tacl_a_00595}.
As demonstrated by previous works, having these datasets in SEA languages will increase robust representation in embeddings.

\noindent
\textbf{Training Algorithm.} 
From our ablation study in Appendix~\ref{appedix:lang_correlation}, we found that there are a lot of false positive and negative occurrences during the testing of the models in Table~\ref{tab:sea-mteb_language_results}.
As shown in Figure~\ref{subfig:Language_correlation_positive_multilingual-e5-large-instruct}, the experimental results from multilingual-e5-large-instruct demonstrate that the similarity of positive pairs is high (more than 0.87 in all cases).
However, the similarity of negative pairs is also high (ranging from 0.74 to 0.81), resulting in the overlap between positive and negative pairs (Figure~\ref{subfig:Language_correlation_negative_multilingual-e5-large-instruct}).
Moreover, when the model performs worst in SEA-BED (multilingual-MiniLM-L12-v2), Figures~\ref{subfig:Language_correlation_positive_multilingual-MiniLM-L12-v2} and \ref{subfig:Language_correlation_negative_multilingual-MiniLM-L12-v2} show that the contrast between positive and negative samples is better than robust models, where this model did not employ contrastive learning, unlike the SOTA model.
This highlights the inconsistency of robust models, which necessitates immediate correction.
To mitigate this, recent works~\cite{limkonchotiwat-etal-2022-congen,li2023angle,wang2024denosent} demonstrate the possibility of contrasting positive and negative samples more effectively than vanilla contrastive learning~\cite{simcse}, which is employed in current models.
However, these techniques have not been well explored on SEA languages; the effects and failure cases need further study.
Applying these techniques can mitigate the issue of overlap.

\noindent
\textbf{Architecture.}
Similar to the findings from previous works in Appendix~\ref{appedix:tokenizer_analysis}, the tokenizer plays a crucial role in embeddings.
In our findings, we discovered that the current models' tokenizer does not include a Telugu token; adding Telugu would enhance the representation of these models.
However, Appendix~\ref{appendix:dataset_analysis} also shows that adding tokens is not effective for all languages; in most cases, non-Latin scripts will have the most effect.
We can also employ adding token techniques~\cite{cui2024efficienteffectivetextencoding,nguyen-etal-2024-seallms} by adding new tokens with minimal effort during continual pre-training to maintain previous knowledge and incorporate new knowledge into the model.

In summary, future work can benefit from the insights gained from our discussions.
The fastest and most cost-effective way is to obtain more datasets using machine translation.
In particular, we can utilize models that demonstrate robust performance for English-to-SEA language translation on SEA translation benchmarks~\cite{susanto2025sea-helm}, such as ChatGPT or Google Gemini.
%
%
Then, we can focus on applying and adapting novel training objectives to SEA embeddings.
We also need to study how to adapt from an English-centric to a SEA-centric approach, leaving a gap for future work to propose a new training objective for the multilingual scenario. 
Lastly, we can focus on the changes in architecture since this will require a new round of pre-training. 
However, not all languages will benefit from this change; we need to carefully add new tokens to the model, as previous studies have shown.

%% file: 2_Related_work.tex
\section{Related Work}
\label{section_related_work}

\subsection{Text Embedding Benchmarks}

Existing text embedding benchmarks primarily focus on high-resource languages. 
Notable examples include SentEval~\cite{conneau2018senteval}, which provides a preliminary benchmark for understanding text embeddings in STS and transfer learning.
USEB~\cite{wang2021useb} is an unsupervised embedding benchmark focusing on pair-text classification.
BEIR~\cite{Thakur2021beir} is a heterogeneous benchmark focusing only on 18 information retrieval datasets.
MTEB~\cite{Muennighoff2023mteb} is a large-scale version of BEIR that not only focuses on retrieval tasks but also on diverse tasks, i.e., bitext mining, classification, and semantic textual similarity. 
However, these benchmarks primarily focus on English, while many works extend MTEB from English to Chinese~\cite{xiao2024cpack}, German~\cite{wehrli-etal-2023-german}, and French~\cite{ciancone2024mtebfrenchresourcesfrenchsentence}. 
Recently, an attempt has been made to create a multilingual version of MTEB, called MMTEB~\cite{enevoldsen2025mmteb}.
MMTEB multilingual benchmark evaluates 10 tasks and 270 datasets; notably, only 22 of these datasets are from SEA languages.
Thus, results from MMTEB might not be representative of performance in SEA languages, given the reliance on machine-translated datasets.

\subsection{SEA Benchmarks} 
There have been many efforts to formulate SEA benchmarks.
NusaCrowd~\cite{cahyawijaya2023nusacrowd} proposed a large-scale Indonesian benchmark focusing on natural language understanding and generation, especially for decoder models. 
VN-MTEB~\cite{pham-etal-2026-vn} a Vietnamese text embedding benchmark with 41 datasets across six tasks, supported by a scalable pipeline using LLM-based translation, semantic filtering, and LLM-as-a-judge for quality assurance in low-resource settings
SEACrowd~\cite{lovenia2024seacrowd} and SEA-VL~\cite{cahyawijaya-etal-2025-crowdsource} are data collection projects that gather SEA benchmarks in their own repositories. 
%
%
The experiment from SEA projects focuses primarily on large language models, particularly the Llama~\cite{dubey2024llama3} and T5~\cite{raffel2020t5} families. 
Moreover, these benchmarks do not accurately measure the effectiveness of embedding in SEA texts.
In particular, previous works studied large language models and generative outputs, while embeddings have not been experimented with in SEA languages.


%% file: 6_Conclusion.tex
\section{Conclusion}

Our analyses show that multilingual embedding performance in SEA languages is highly conditional, varying across models, tasks, and language-task combinations. 
Language-model comparisons reveal that no single model consistently performs well across all languages, with substantial disparities persisting even among the strongest models. 
Task-model analyses reveal clear differences in task difficulty: while classification and bitext mining approach saturation for well-resourced languages, clustering and semantic similarity remain challenging and unstable. 
Language-task comparisons demonstrate uneven performance within individual languages, showing that success on one task does not reliably generalize to others.


Our results suggest that observed performance gaps arise from interrelated limitations in data coverage, training objectives, and architectural design. 
\textbf{Dataset availability} plays a central role: models trained predominantly on English-centric or weakly aligned multilingual data struggle to generalize across languages and tasks. 
\textbf{Training algorithms} need to address the high positive-negative similarity overlap in low-resource languages, suggesting the application of cross-lingual transfer to ensure that task-relevant semantic structures learned in one language generalize to others. 
\textbf{Architectural factors}, including tokenizer design and language coverage, introduce additional structural constraints that disproportionately affect non-Latin and underrepresented languages. 
Consequently, improving multilingual embeddings for SEA languages requires coordinated advances across datasets, training algorithms, and architectures.

\subsection*{Acknowledgement}

This research is supported by the National Research Foundation, Singapore, under its National Large Language Models Funding Initiative. Any opinions, findings, and conclusions or recommendations expressed in this material are those of the author(s) and do not reflect the views of the National Research Foundation, Singapore.

\section*{Limitations}

While SEA-BED substantially expands the landscape of multilingual sentence-embedding evaluation, several limitations remain.

First, the coverage is uneven across the 10 SEA languages. 
Although the benchmark encompasses the region's major language families and scripts, some languages are represented in fewer task categories due to the limited availability of publicly accessible datasets. 
For extremely low-resource languages such as Tetum, the limited availability of high-quality datasets undermines the reliability of evaluation results, necessitating careful interpretation and preventing definitive conclusions.
This asymmetry limits the breadth of task-language combinations that can be examined uniformly.

Second, the scope of evaluation data is regionally bounded. 
SEA-BED focuses on locally grounded and native-verified data from Southeast Asian linguistic communities. 
While this enables strong internal validity for SEA-specific probing, it does not capture cultural or pragmatic phenomena unique to other low-resource regions, nor does it fully represent global diversity.

Finally, the benchmark incorporates machine-generated and machine-derived data. While this supports broader task coverage and scalability, such data may differ systematically from human-authored text. 
Given persistent coverage constraints, the inclusion of machine-derived data is often unavoidable in multilingual evaluation. 
Accordingly, we distinguish evaluation results by data source where applicable to examine differences between human-authored and machine-derived conditions in Appendix~\ref{appendix:translation_vs_human}. 
However, we do not conduct controlled studies isolating the causal effects of machine generation, nor do we claim that machine-derived data uniformly approximates human-authored data across tasks.

\section*{Ethical Statement}
For the annotator details, we hired annotators (graduated students) who speak SEA languages natively (see Appendix~\ref{appendix:annotator_demographics} for more details). 
We first ran the annotation experiment and selected only the annotators who passed the annotation test, i.e., the English test and NLP understanding, to test whether annotators understand and can perform work in a high-quality manner.
In addition, the payment rate for each annotator is 18 USD/Hr, which is considered higher than the average payment. 

%% file: 7_Appendix.tex
\appendix

\onecolumn
\section*{Appendix}



\section{Annotator Demographics and Guidelines}
\label{appendix:annotator_demographics}

\noindent
\textbf{Data Assurance Annotators}
We hired two native speakers of each SEA language to check the quality of the datasets before adding them to SEA-BED.
These annotators are graduate students who have passed the English test and the NLP test (i.e., understanding the concepts of each task in SEA-BED).
We give them the guidelines as follows.
\emph{Please re-check the datasets that (i) the correctness of text and written style is natural and understandable for a native speaker; (ii) the correctness of the gold label, e.g., a correct class of text classification is assigned.}
We have asked them to check 182 datasets.
There are some datasets that utilize machine translation with low-quality outputs or poor-quality labels; we remove them from our SEA-BED.

\noindent
\textbf{New Dataset Annotators}
In this work, our collaborators helped us translate the data from English to Thai and Burmese for STS and NLI tasks.
These people are Thai and Burmese undergraduate and graduate students studying in Thailand, aged from 20 to 25 years old, who can speak English and their native language (Thai or Burmese).
We use three Thai annotators and one Burmese annotator to create new datasets.
We also removed some examples that contain special characters that cannot be shown in Google Sheets. 
We give them the guidelines as follows.
\emph{Translate the selected datasets to make them a human-like or everyday conversation in your native languages} and \emph{change the subject of a sentence to be gender-neutral} since both the Thai and Burmese languages have words or morphemes that can express the gender of the speaker. 
Therefore, the quality of our new human-crafted dataset is higher than that of using machine translations or LLMs to generate data, as such methods have been observed to be less native-like or unrepresentative of natural language use~\cite{lovenia2024seacrowd,singh2025globalmmluunderstandingaddressing}.

\begin{table*}[h!]
\vspace{-1mm}
\centering
\setlength\doublerulesep{4pt}
\scalebox{0.9}{
    \renewcommand{\arraystretch}{1.2}
    \begin{tabular}{lcc}
        \toprule
        \textbf{ISO Language Name } & \textbf{ISO 639-3} & \textbf{Number of speakers} 
        \\
        \bottomrule
        \midrule
        Indonesian & ind  & $\sim$ 200 million \\
        Thai & tha & $\sim$ 60 million  \\
        Vietnamese & vie & $\sim$ 85 million \\
        Burmese & mya & $\sim$ 43 million \\
        Filipino & fil & $\sim$ 45 million \\
        Khmer & khm & $\sim$ 17 million \\
        Malay & zsm & $\sim$ 33 million  \\
        Lao & lao & $\sim$ 7 million \\
        Tamil & tam & $\sim$ 85 million \\
        Tetum & tet & $\sim$ 1.3 million \\
        \bottomrule
    \end{tabular}
    }
    \vspace{-2mm}
    \caption{Overview of Southeast Asian languages, including ISO language names, ISO 639-3 codes, and approximate numbers of speakers (L1 and L2 combined), based on Ethnologue (2023/2024).}
    \vspace{-3mm}
    \label{tab:number_of_speaker}
\end{table*}

\section{Benchmark Efficiency} \label{appendix:benchmark_efficiency}

\noindent
\textbf{Caching Embeddings}. To improve the run-time efficiency, we use embedding caching to store embedded texts in memory and cache files; when seen texts are input to the same model, we will use the cached embedding instead of computing the new one to decrease the run-time of our benchmark.

\noindent
\textbf{Downsampling.}
\citet{enevoldsen2025mmteb} proposed a downsampling technique for the English benchmark, decreasing the number of samples by 98\%.
However, as shown in Table~\ref{tab:down_sampling}, we applied the same technique to our benchmark (bitext mining datasets) and found that the performance of each model increased in all cases.
This is because all challenging samples may have been removed from the dataset, leading to improved performance for most models.
%
%
Moreover, the ranking of each model changed, in contrast to the findings of \citet{enevoldsen2025mmteb}, where the rankings remained largely unchanged.
Therefore, we did not apply the downsampling technique to our benchmark.

\begin{table*}[h!]
\vspace{-1mm}
\centering
\setlength\doublerulesep{4pt}
\scalebox{0.7}{
    \renewcommand{\arraystretch}{1.2}
    \begin{tabular}{lccc}
        \toprule
        \textbf{Model} & \textbf{100\% Dataset} & \textbf{30\% Dataset} & \textbf{Rank after downsampling} \\
        \bottomrule
        \midrule
        multilingual-e5-large-instruct (560M) & 87.86 & 93.03 & 0 \\
        Qwen3-Embedding-8B (8B) & 84.78 & 90.31 & \color[HTML]{FE0000}$\downarrow$1 \\
        bge-multilingual-gemma2 (9B) & 82.02 & 90.71 & \color[HTML]{008000}$\uparrow$3 \\
        multilingual-e5-large (560M) & 84.51 & 88.19 & \color[HTML]{FE0000}$\downarrow$1 \\
        bge-m3 (568M) & 86.18 & 91.89 & \color[HTML]{008000}$\uparrow$1 \\
        GritLM-7B (7B) & 63.63 & 69.68 & 0 \\
        e5-mistral-7b-instruct (7B) & 65.30 & 73.42 & 0 \\
        Qwen3-Embedding-0.6B (595M) & 56.53 & 62.95 & 0 \\
        multilingual-mpnet-base (278M) & 68.12 & 73.97 & 0 \\
        LaBSE (471M) & 86.84 & 90.51 & \color[HTML]{FE0000}$\downarrow$2 \\
        multilingual-MiniLM-L12 (118M) & 53.23 & 59.06 & 0 \\
        Gemma-SEA-LION-v3-9B-IT (9B) & 15.31 & 3.21 & \color[HTML]{FE0000}$\downarrow$1 \\
        Sailor2-8B-Chat (8B) & 4.31 & 6.01 & \color[HTML]{008000}$\uparrow$1 \\
        \bottomrule
    \end{tabular}
    }
    \vspace{-2mm}
    \caption{We evaluate 13 models on bitext mining using 100\% and 30\% dataset sizes. We also indicate the rank change of the model before and after downsampling to show the performance discrepancy.}
    \vspace{-3mm}
    \label{tab:down_sampling}
\end{table*}

\section{Domains}
\label{appendix:domain_details}
For domains in the SEA-BED benchmark, we include the following:
\begin{compactitem} 
    \item \textbf{Academic:} Formal writing and research publications commonly found in scholarly journals, theses, and dissertations.
    \item \textbf{Blog:} Informal, conversational writings about a variety of topics published on websites or personal blogs.
    \item \textbf{Constructed:} Artificially created text or speech, often in experiments to target particular abilities. 
    \item \textbf{Encyclopedic:} Structured, reference-based texts offering thorough and factual information on various topics.
    \item \textbf{Fiction:} Narrative writing that involves creative content, such as novels, short stories, and other storytelling forms.
    \item \textbf{Government:} Documents, reports, and publications officially issued by government agencies.
    \item \textbf{Legal:} Documents and texts concerning laws, legal processes, contracts, and legal theories.
    \item \textbf{Medical:} Scientific and clinical publications focused on healthcare, treatments,  patient care, and medical studies.
    \item \textbf{News:} News articles and reports that address current events, political developments, economic trends, and other timely topics.
    \item \textbf{Non-fiction:} Texts grounded in real events and factual information, including biographies, essays, and documentaries.
    \item \textbf{Religious:} Writings concerning religious teachings, doctrines, sacred texts, and discussions on spirituality.
    \item \textbf{Reviews:} Analytical assessments of books, films, music, products, or services.
    \item \textbf{Social:} Messages and conversations shared on social media, online forums, and other digital platforms.
    \item \textbf{Spoken:} Spoken content such as speeches, dialogues, interviews, and recorded discussions.
    \item \textbf{Subtitles:} Written transcriptions or translations of spoken content from films, videos, or multimedia presentations.
    \item \textbf{Web:} Web-based content spanning diverse topics, often featuring hyperlinks and multimedia elements.
    \item \textbf{Written:} A broad category encompassing all forms of text-based communication, both print and digital.
\end{compactitem}

\section{Performance Changes Analysis} \label{appedix:ranking_breakdown}
Here, we are examining how SEA-focused performance contrasts with the broader multilingual benchmark (MMTEB). 
To study the robustness of embeddings in world and SEA languages, we compare the ranking changes between our benchmark and the multilingual text embedding benchmark, MMTEB.
We use the task average metric (Table~\ref{tab:sea-mteb_task_results}), similar to MMTEB.

\begin{figure}[h!]
\hspace{-1mm}
\vspace{-4mm}
\centering
\centering
\includegraphics[width=0.48\textwidth]{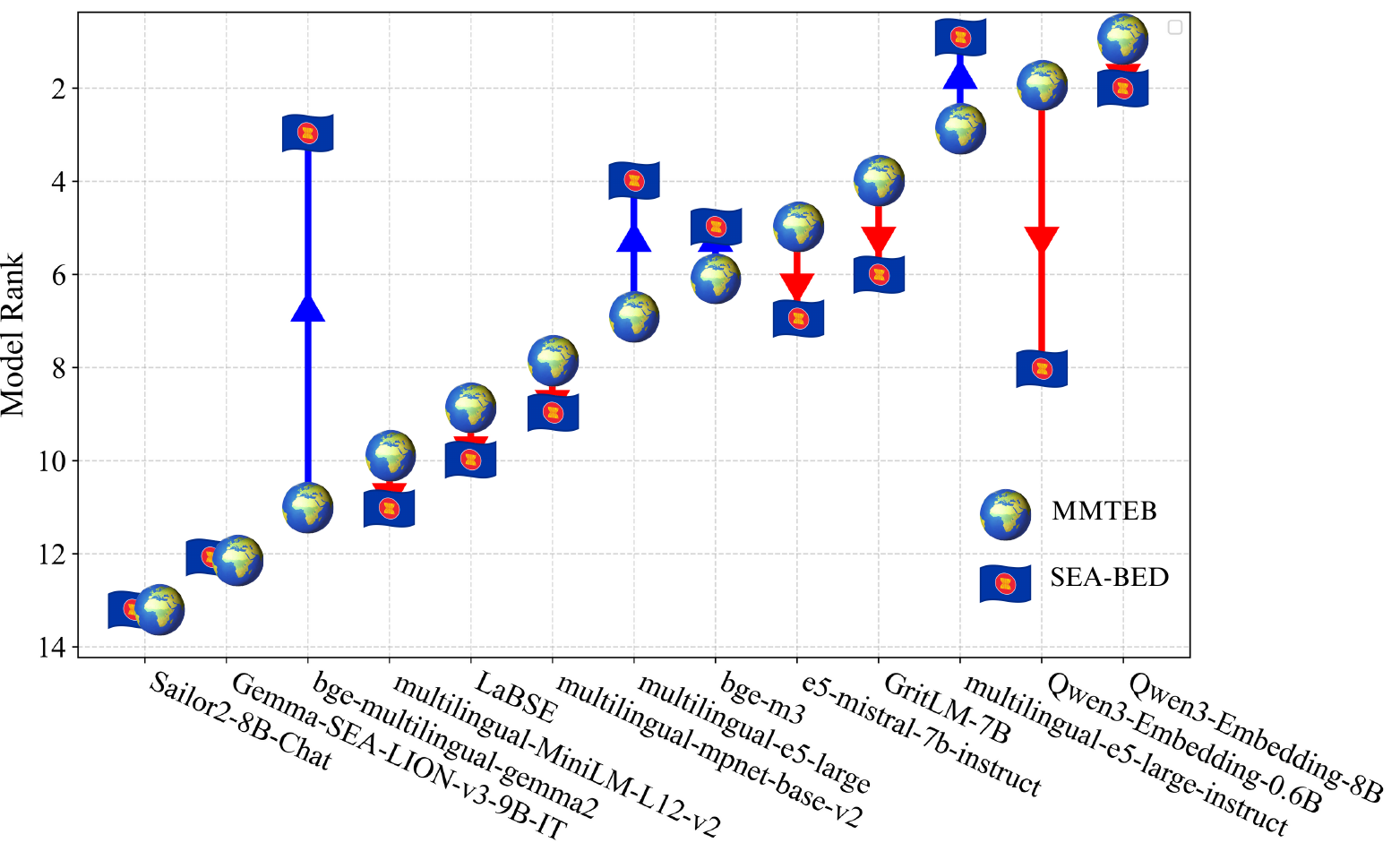}
\caption{Ranking difference between MMTEB and SEA-BED.}
\vspace{-6mm}
\label{fig:model_rank_mmteb_vs_sea-mteb}
\end{figure}

As shown in Figure~\ref{fig:model_rank_mmteb_vs_sea-mteb}, based on the experiment from MMTEB, Qwen3-Embedding-8B performed the best on world results, which includes 1,090 languages\footnote{We obtained the model rankings on Nov 28th, 2025.}.
However, when we focus only on SEA languages using SEA-BED, the ranking of Qwen3-Embedding-8B dropped from first to second place.
In addition, Qwen3-Embedding-0.6B dropped from second rank to eighth rank.
This is because some of the linguistic and dialect knowledge will be different compared to other groups of languages, when we evaluate them only for the SEA languages. 
%
%
%
This highlights that the challenges, gaps, and model capabilities measured in MMTEB and our benchmark differ, particularly in the supported languages for embedding models that do not fully support SEA languages.
Although some models are effective in performing well on MMTEB, they are not guaranteed to achieve the same performance for SEA languages.
%

\section{Tokenizer Analysis} \label{appedix:tokenizer_analysis} 

This section examines whether limited coverage of SEA vocabulary in multilingual tokenizers correlates with poor downstream results. 
We highlight scripts like Lao or Khmer, which are often underrepresented in tokenizers.
Previous works~\cite{ali2024tokenizerchoice, Arnett2025, Liang2023XLM-V} demonstrated that vocabularies in a tokenizer affect the model performance in downstream tasks.
In particular, when the multilingual tokenizer represents more vocabulary in some languages, the performance on those languages has also been observed to improve.
In this study, we want to investigate whether the vocabulary in the tokenizer affects SEA-BED's overall performance or not. 
To answer this question, we count the SEA tokens in each text embedding model and compare their performance from Table~\ref{tab:sea-mteb_language_results}.

As shown in Table~\ref{tab:number_of_vocab}, the language with the most tokens represented in a tokenizer is Filipino, with an average of 2.94 percent of vocabulary tokens in 13 models.
However, compared to the language performance (Table~\ref{tab:sea-mteb_language_results}), Filipino performance is lower than Indonesian.
Surprisingly, there are no tokens for Tetum at all in the 13 models. 
We observe that performance on Tetum is also the worst compared to other SEA languages. 
Moreover, the performance is mixed for languages that do not use Latin characters, i.e., Thai, Burmese, Lao, and Tamil.

\begin{table*}[h]
    \centering
    \scriptsize
    \renewcommand{\arraystretch}{1.4}
    \begin{tabular}{lcccccccccc}
        \toprule
        \textbf{Model} & \textbf{ind} & \textbf{tha} & \textbf{vie} & \textbf{mya} & \textbf{fil} & \textbf{khm} & \textbf{zsm} & \textbf{lao} & \textbf{tam} & \textbf{tet} \\
        \midrule 
        multilingual-e5-large-instruct (560M) & 1.20 & 1.61 & 0.73 & 0.91 & 3.59 & 0.66 & 0.20 & 0.56 & 0.98 & 0.00 \\  
        Qwen3-Embedding-8B (8B) & 0.39 & 1.70 & 0.84 & 0.02 & 1.13 & 0.03 & 0.11 & 0.02 & 0.02 & 0.00 \\  
        bge-multilingual-gemma2 (9B) & 0.59 & 0.50 & 0.55 & 0.45 & 3.04 & 0.03 & 0.11 & 0.02 & 0.13 & 0.00 \\  
        multilingual-e5-large (560M) & 1.20 & 1.61 & 0.73 & 0.91 & 3.59 & 0.66 & 0.20 & 0.56 & 0.98 & 0.00 \\
        bge-m3 (568M) & 1.20 & 1.61 & 0.73 & 0.91 & 3.59 & 0.66 & 0.20 & 0.56 & 0.98 & 0.00 \\
        GritLM-7B (7B) & 0.27 & 0.19 & 0.55 & 0.45 & 3.04 & 0.03 & 0.11 & 0.02 & 0.13 & 0.00 \\
        multilingual-mpnet-base (278M) & 1.20 & 1.61 & 0.73 & 0.91 & 3.59 & 0.66 & 0.20 & 0.56 & 0.98 & 0.00 \\ 
        LaBSE (471M) & 1.12 & 0.45 & 0.81 & 0.45 & 4.65 & 0.54 & 0.19 & 0.29 & 1.28 & 0.00 \\ 
        e5-mistral-7b-instruct (7B) & 0.27 & 0.19 & 0.55 & 0.45 & 3.04 & 0.03 & 0.11 & 0.02 & 0.13 & 0.00 \\
        Qwen3-Embedding-0.6B (595M) & 0.39 & 1.70 & 0.84 & 0.02 & 1.13 & 0.03 & 0.11 & 0.02 & 0.02 & 0.00 \\
        multilingual-MiniLM-L12 (118M) & 1.20 & 1.61 & 0.73 & 0.91 & 3.59 & 0.66 & 0.20 & 0.56 & 0.98 & 0.00 \\  
        Gemma-SEA-LION-v3-9B-IT (9B) & 0.59 & 0.50 & 0.55 & 0.45 & 3.04 & 0.03 & 0.11 & 0.02 & 0.13 & 0.00 \\
        Sailor2-8B-Chat (8B) & 0.39 & 1.70 & 0.84 & 0.02 & 1.13 & 0.03 & 0.11 & 0.02 & 0.02 & 0.00 \\
        \midrule
        Average & 0.77 & 1.15 & 0.71 & 0.53 & 2.94 & 0.31 & 0.15 & 0.25 & 0.52 & 0.00 \\
        \midrule
    \end{tabular}
    \vspace{-2mm}
    \caption{The percentage number of vocabulary tokens for each model in each language.}
    \vspace{-2mm}
    \label{tab:number_of_vocab}
\end{table*}

\section{Language Similarity} \label{appedix:lang_correlation}

To further understand the similarity between language and performance, we analyze the performance of bi-text retrieval datasets in SEA languages.
In particular, we study the language similarity of robust and non-robust models, e.g., top-performing and worst-performing embedding models, to see what the desired property is to improve our benchmark.
We utilize the dialect pairing subset task in this experiment, where we use a batch size of 128 for the negative pair evaluation. 
In addition, we use cosine similarity as the main metric, where higher values indicate greater embedding similarity between language pairs.

\begin{figure*}[h!]
\centering
\begin{subfigure}[b]{0.49\textwidth}
\vspace{-4mm}
\centering
\includegraphics[width=0.8\textwidth]{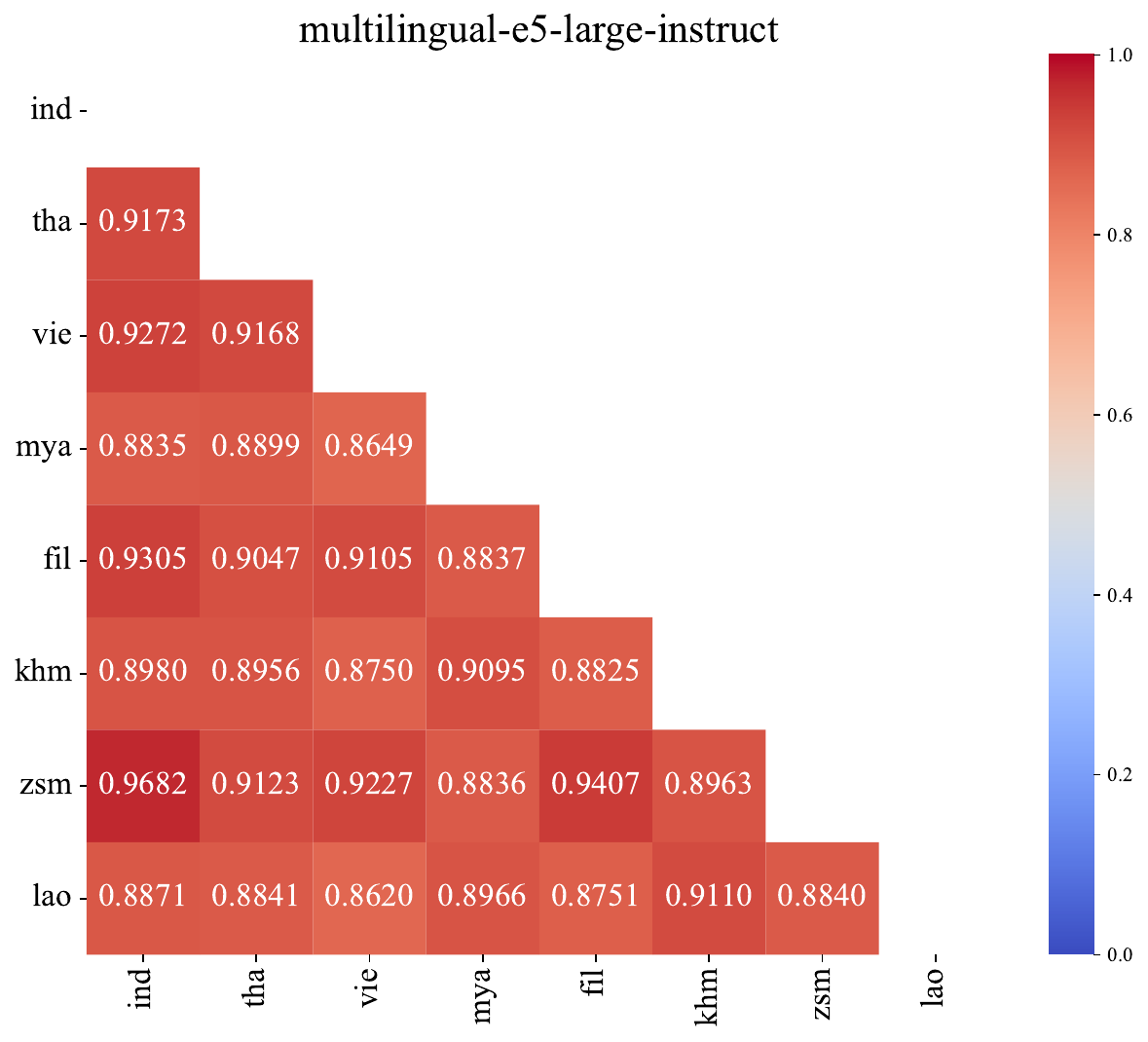}
\caption{The top-performing model on the positive pairs}
\label{subfig:Language_correlation_positive_multilingual-e5-large-instruct}
\end{subfigure}
\hfill
\begin{subfigure}[b]{0.49\textwidth}
\centering
\includegraphics[width=0.8\textwidth]{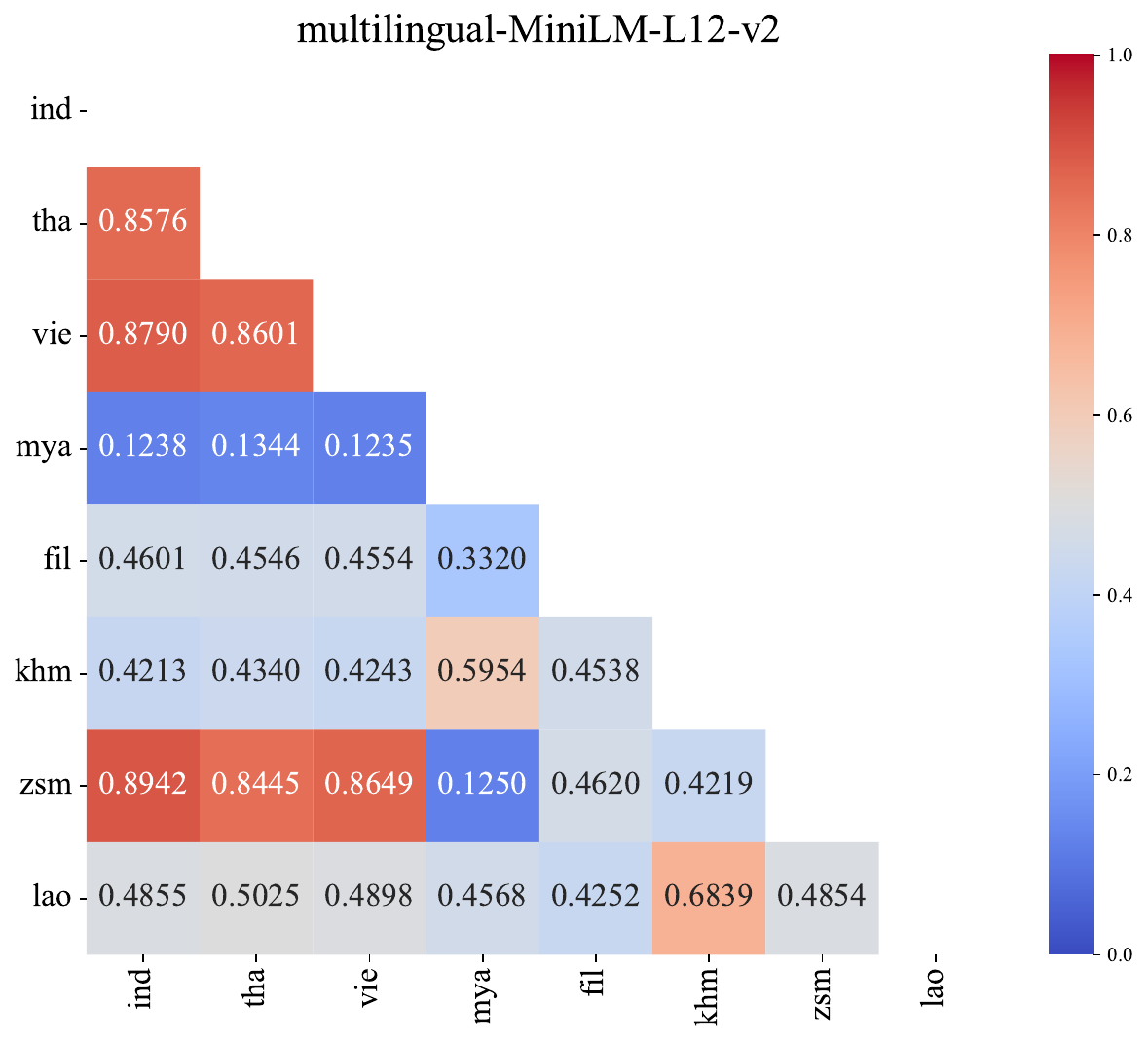}
\caption{The worst-performing model on the positive pairs}
\label{subfig:Language_correlation_positive_multilingual-MiniLM-L12-v2}
\end{subfigure}
\hfill
\begin{subfigure}[b]{0.49\textwidth}
\centering
\includegraphics[width=0.8\textwidth]{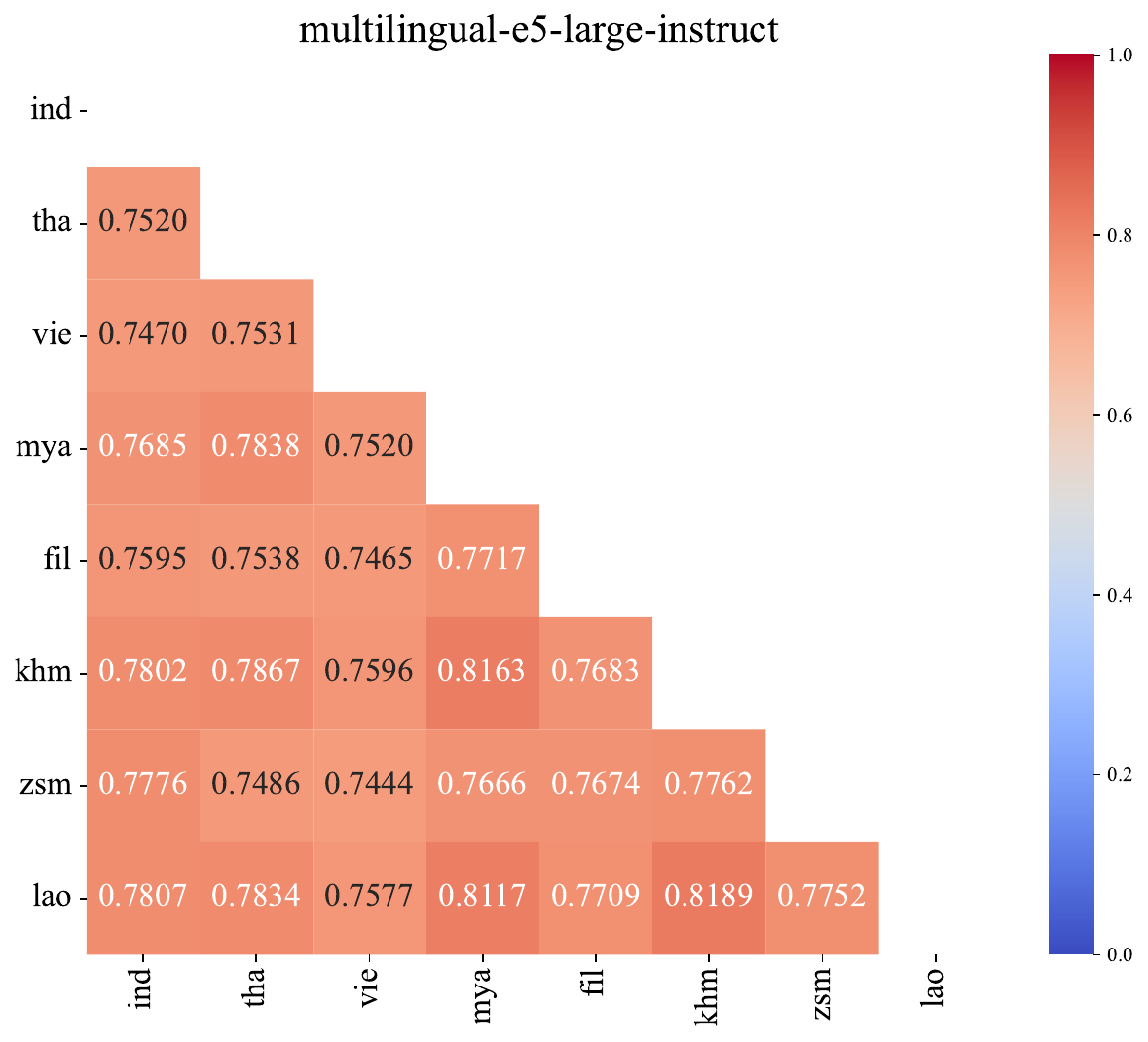}
\caption{The top-performing model on the negative pairs}
\label{subfig:Language_correlation_negative_multilingual-e5-large-instruct}
\end{subfigure}
\hfill
\begin{subfigure}[b]{0.49\textwidth}
\centering
\includegraphics[width=0.8\textwidth]{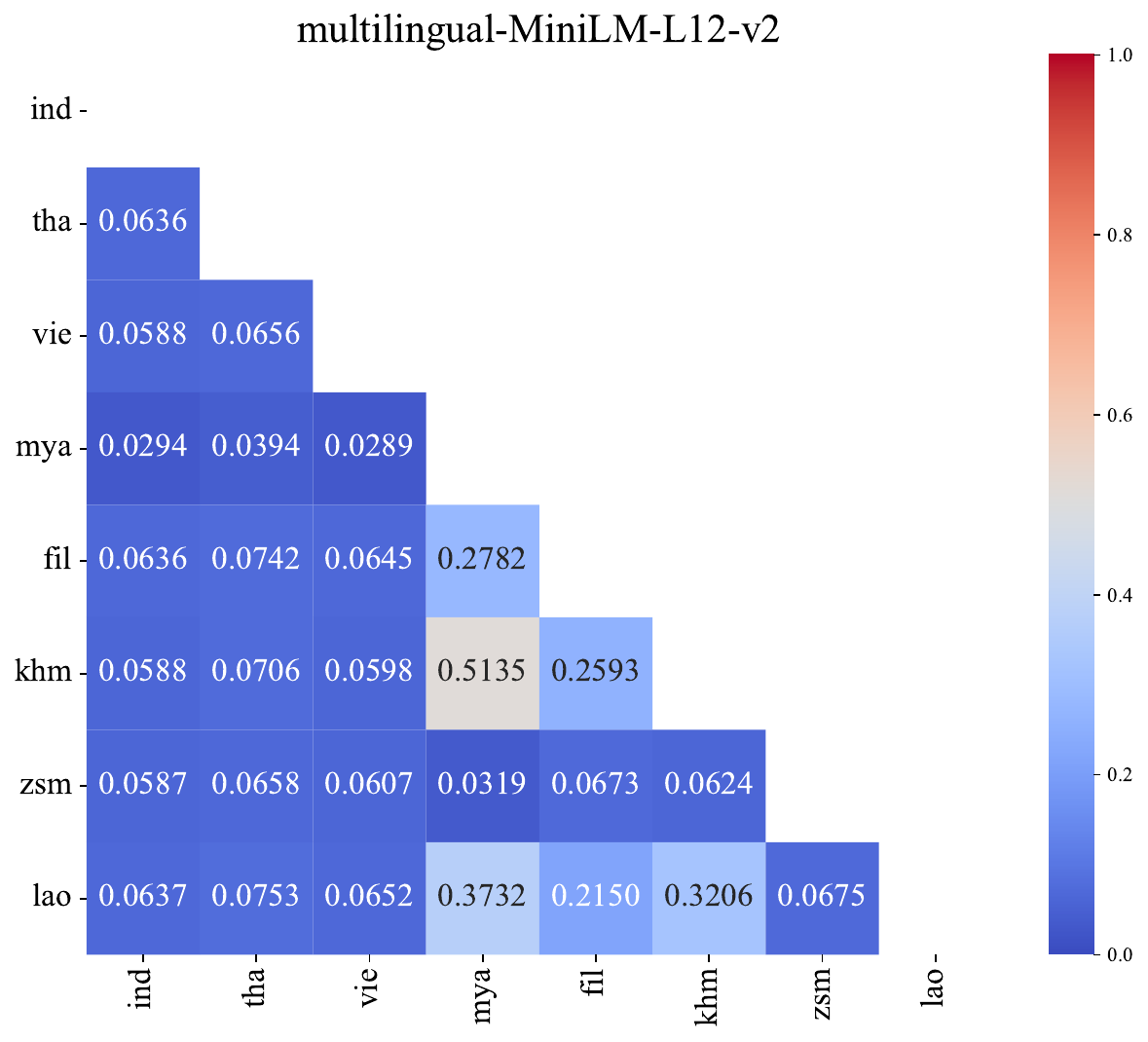}
\caption{The worst-performing model on the negative pairs}
\label{subfig:Language_correlation_negative_multilingual-MiniLM-L12-v2}
\end{subfigure}
\caption{We perform cross-lingual similarity using the bitext mining task (dialect pairing subset). 
(Top) Cross-lingual similarity metrics of the top-performing and worst-performing embedding models on the \underline{positive} parallel samples.
(Bottom) Cross-lingual correlation metrics of the top-performing and worst-performing embedding models on the \underline{negative} parallel samples.
}
\vspace{-1mm}
\label{fig:language_correlation}
\end{figure*}

As shown in Figure~\ref{fig:language_correlation}, the top-performing model, multilingual-e5-large-instruct, shows consistently high similarity for positive samples, especially Indonesian-Malay (0.9682 points), Indonesian-Filipino (0.9305 points), and Thai-Vietnamese (0.9168 points), indicating strong cross-lingual embeddings. 
However, multilingual-e5-large-instruct unexpectedly maintains high similarity for negative samples (0.75-0.81 points), indicating limited distinction between unrelated sentence pairs and highlighting a gap for improvement.
In contrast, multilingual-MiniLM-L12-v2 struggles with related positive pairs, showing lower similarity for Indonesian-Filipino (0.4601 points) and notably weak similarity with Burmese (around 0.12-0.59 points).
Interestingly, this model achieves low similarity for negative pairs, mostly under 0.08 points, clearly distinguishing unrelated samples.
Although it falls short in overall embedding quality, multilingual-MiniLM-L12-v2's distinct negative sample separation provides valuable insights into desirable characteristics for embedding models.
These findings suggest that a balanced approach, achieving both strong cross-lingual similarity for positive examples and clear differentiation for negative examples, is essential to improve future embedding benchmarks.

\section{Machine vs. Human Datasets} \label{appendix:translation_vs_human} 

This section testing whether human-crafted data yields results different from machine-generated data.
We split the experiment into machine generation and translation studies.

\begin{figure}[h!]
\centering
\includegraphics[width=0.45\textwidth]{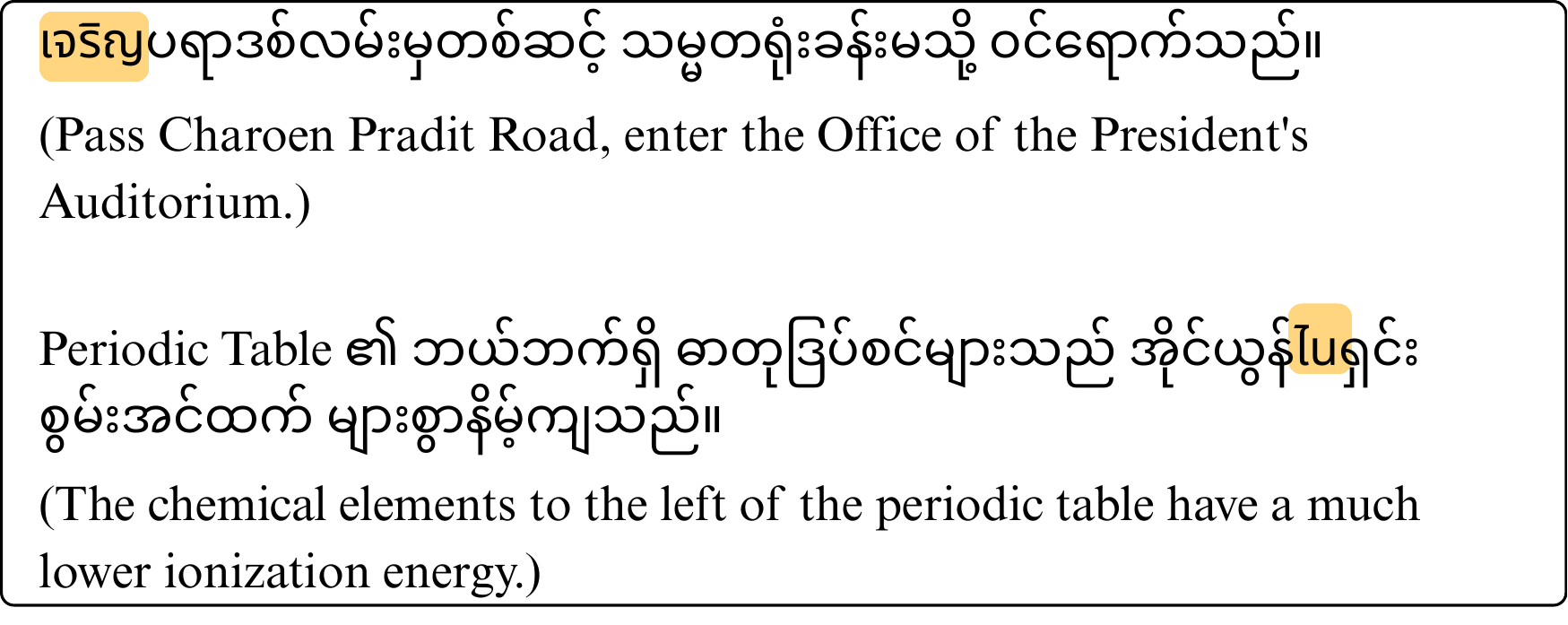}
\vspace{-3mm}
\caption{Code-switching between Thai and Burmese words (translated by Google NMT).}
\vspace{-3mm}
\label{fig:incorrect_character}
\end{figure}

\noindent
\textbf{Machine Translation vs. Human-translated Datasets.}
To compare machine-translated and human-translated data, we evaluate our new Thai and Burmese STS datasets from Table~\ref{tab:new_dataset_statistic} against versions translated by Google’s MT system. 
As shown in Table~\ref{tab:translation_vs_human}, Thai results differ by less than 2 Spearman points across all settings, aligning with prior work showing that English-Thai NMT is already reliable for practical use~\cite{DBLP:journals/lre/LowphansirikulP22,chiaranaipanich2024generalpurposelargelanguagemodels}.
In contrast to Thai, the performance gap between Burmese human and machine translation datasets is larger than that of Thai in most cases.
We found that the Google NMT results for Burmese sometimes show code-switching between Thai and Burmese characters, as shown in Figure~\ref{fig:incorrect_character}.
This emphasizes that, in underrepresented languages, using humans to create evaluation datasets is still better than relying on machine translations.

\noindent
\textbf{Machine Generation vs. Human Datasets.}
While many recent works introduce datasets generated with machine learning for scalability, we argue that fully machine-generated data remains unstable and should not dominate benchmarks, as it can distort model performance and research conclusions.
To illustrate this, we compare human-crafted and machine-generated datasets using the top five models from our previous study. Keeping the same tasks and languages, we evaluate only datasets created by either humans or machines. 
The results for both settings are reported in Tables~\ref{tab:sea-bed-language-results}, \ref{tab:sea-bed-task-results} and~\ref{tab:sea-bed-language-task-results}.

Our results reveal two main effects: (i) shifts in average performance and (ii) shifts in model ranking. 
First, machine-generated datasets almost always reduce performance relative to human-crafted ones, with the sole exception of bge-multilingual-gemma2. This finding aligns with our results for machine-translated data (Table~\ref{tab:translation_vs_human}), indicating that machine-generated inputs can degrade model performance.
Second, rankings become unstable, making evaluations unreliable. A robust benchmark should align with human-based outcomes, yet machine-generated datasets fail to preserve ranking consistency. 
Tetum offers a clear example: in Table~\ref{tab:sea-bed-language-results}, bge-multilingual-gemma2 leaps from 30.91 to 99.18 points. This occurs because the Tetum machine-generated set resembles a simple language-detection task from MADLAD-400, where all models score above 90, suggesting data leakage or in-domain overlap.
Additional analysis of the machine-generated datasets is provided in Appendix~\ref{appendix:dataset_analysis}.

\begin{table*}[ht!]
\centering
\setlength\doublerulesep{4pt}
\scalebox{0.55}{
    \renewcommand{\arraystretch}{1.2}
    \begin{tabular}{lc|ccc|ccc}
        \toprule
        \textbf{Model} & \textbf{Original (eng)} & \textbf{Mach. (mya)} & \textbf{Hum. (mya)} & \textbf{Diff. (mya)} & \textbf{Mach. (tha)} & \textbf{Hum. (tha)} & \textbf{Diff. (tha)} \\
        \bottomrule
        \midrule
        multilingual-e5-large-instruct (560M) & 82.87 & 74.82 & 75.06 & 0.24 & 79.66 & 79.80 & 0.14  \\
        Qwen3-Embedding-8B  (595M) & 81.17 & 74.02 & 75.81 & 1.79 & 80.75 & 80.74 & 0.01 \\
        bge-multilingual-gemma2 (9B) & 84.64 & 75.51 & 72.87& 2.64 & 80.25 & 78.97 & 1.28 \\
        multilingual-e5-large (560M) & 80.00 & 71.49 & 71.55& 0.06 & 76.44 & 76.50 & 0.06 \\
        bge-m3 (568M) & 80.86 & 74.57 & 71.96 & 2.61 & 77.75 & 76.07 & 1.68 \\
        GritLM-7B (7B) & 82.65 & 65.60 & 66.03 & 0.43 & 74.64 & 74.81 & 0.17 \\
        e5-mistral-7b-instruct (7B) & 81.86 & 62.36 & 64.63 & 2.27 & 74.57 & 74.57 & 0.00 \\
        Qwen3-Embedding-0.6B  (8B) & 80.11 & 67.10 & 69.23 & 2.13 & 77.88 & 77.79 & 0.09 \\
        multilingual-mpnet-base (278M) & 80.54 & 72.34 & 71.16 & 1.18 & 72.61 & 72.60 & 0.01 \\
        LaBSE (471M) & 73.50 & 69.06 & 70.04 & 0.98 & 69.29 & 68.83 & 0.46 \\
        multilingual-MiniLM-L12 (118M) & 78.89 & 69.27 & 67.26 & 2.01 & 72.25 & 72.23 & 0.02 \\
        Gemma-SEA-LION-v3-9B-IT (9B)   & 60.42 & 46.50 & 49.29 & 2.79 & 55.97 & 56.01 & 0.04 \\
        Sailor2-8B-Chat (8B)   & 57.94 & 48.79 & 52.89 & 4.1 & 55.85 & 54.36 & 1.49 \\
        \midrule
        \emph{Proprietary models} \\ \midrule
        embed-multilingual-v3.0   & 82.62 & 74.56 & 73.92 & 0.64 & 78.40 & 78.63 & 0.23 \\
        jina-embeddings-v3   & 78.37 & 75.92 & 75.61 & 0.31 & 76.40 & 76.36 & 0.04 \\
        voyage-3   & 81.77 & 69.54 & 68.20 & 1.34 & 77.42 & 73.64 & 3.78 \\
        text-embedding-3-small   & 82.37 & 54.86 & 55.65 & 0.79 & 66.47 & 66.45 & 0.02 \\
        \midrule \bottomrule 
    \end{tabular}
    }
    \vspace{-3mm}
    \caption{Model performance on Machine Translation vs. Human Datasets on our STS datasets.}
    \vspace{-5mm}
    \label{tab:translation_vs_human}
\end{table*}

\section{Dataset Analysis} \label{appendix:dataset_analysis}

\begin{figure*}[h!]
\centering
\begin{subfigure}[b]{0.49\textwidth}
\vspace{-4mm}
\centering
\includegraphics[width=0.8\textwidth]{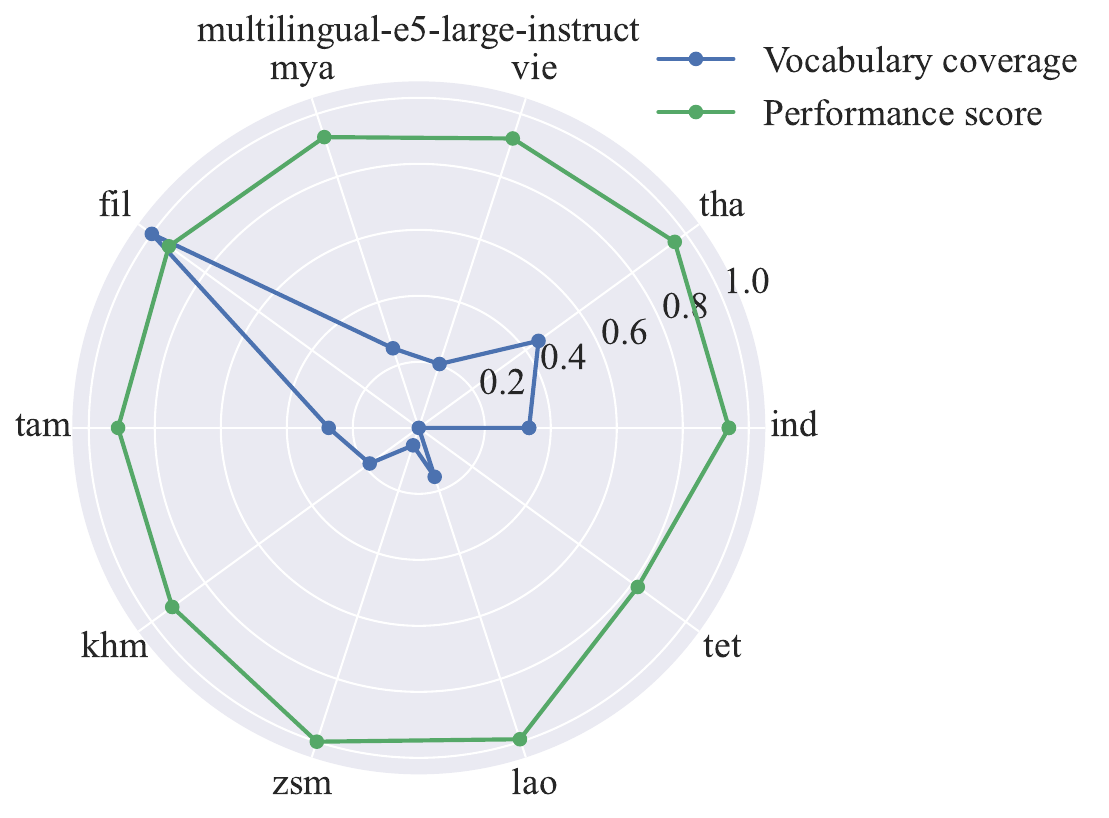}
\caption{}
\label{subfig:multilingual-e5-large-instruct_performance_vs_vocab}
\end{subfigure}
\hfill
\begin{subfigure}[b]{0.49\textwidth}
\centering
\includegraphics[width=0.8\textwidth]{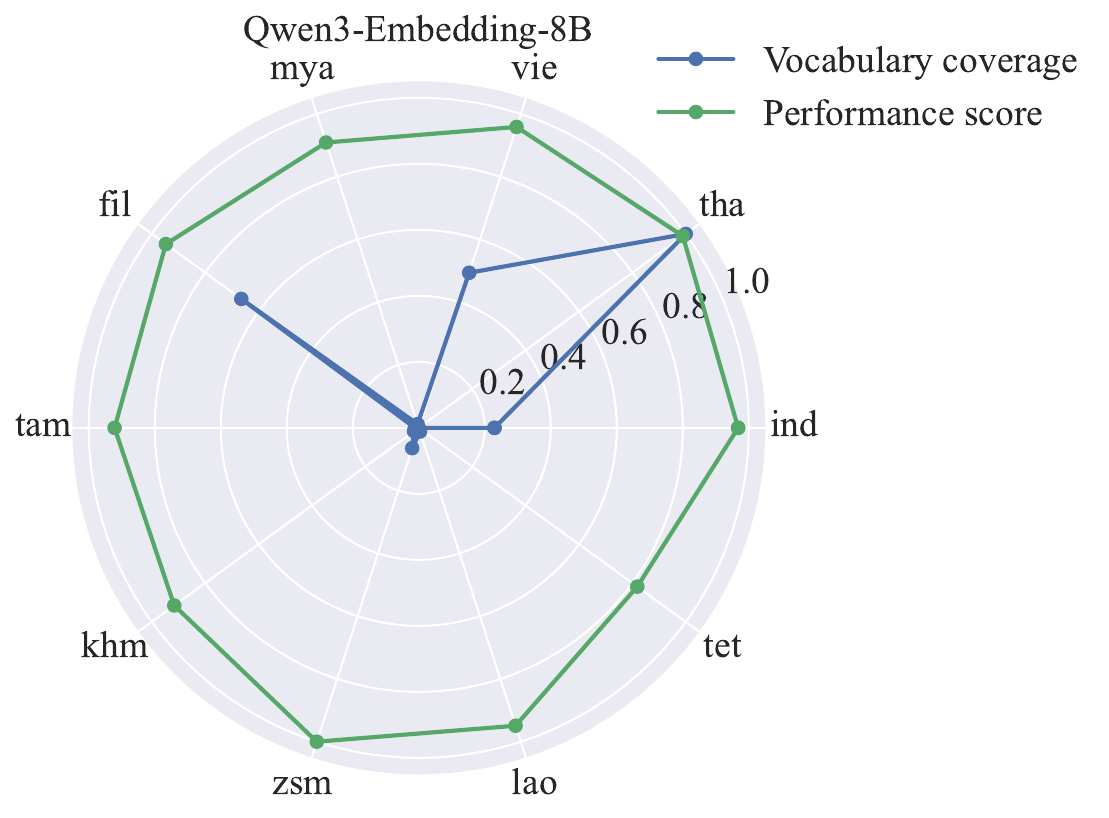}
\caption{}
\label{subfig:Qwen3-Embedding-8B_performance_vs_vocab}
\end{subfigure}
\hfill
\begin{subfigure}[b]{0.49\textwidth}
\centering
\includegraphics[width=0.8\textwidth]{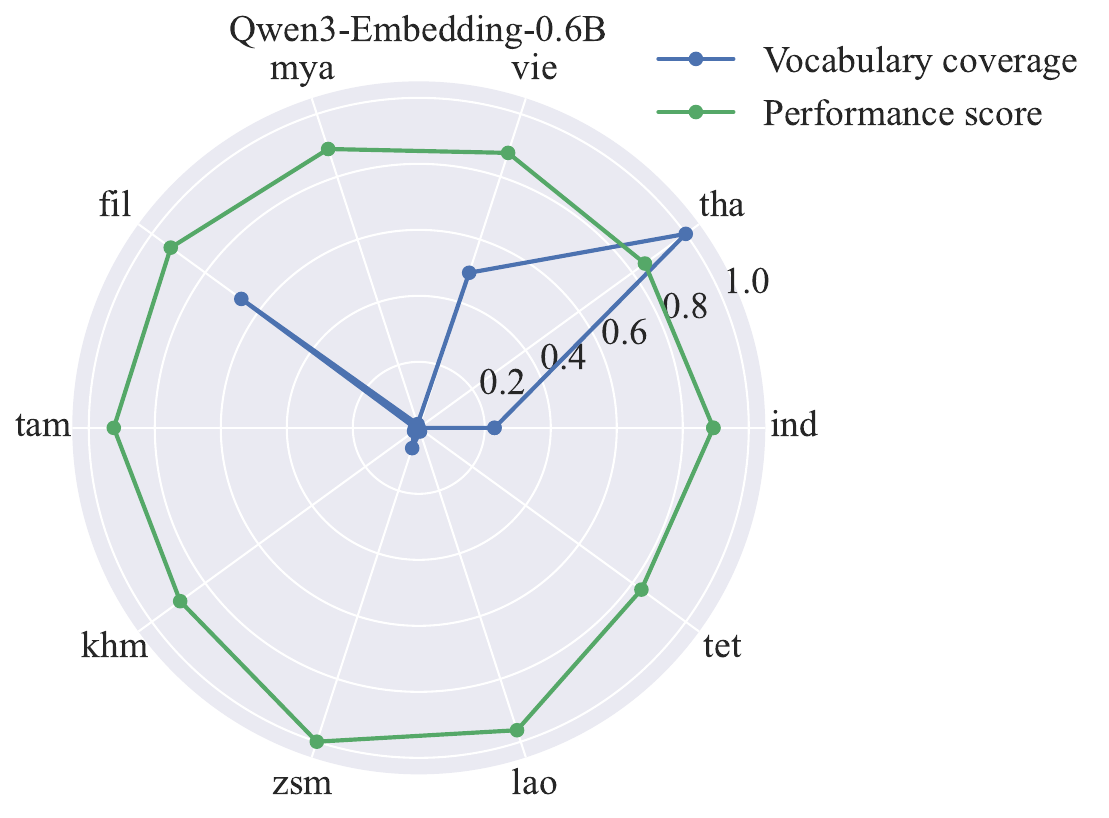}
\caption{}
\label{subfig:Qwen3-Embedding-0.6B_performance_vs_vocab}
\end{subfigure}
\hfill
\begin{subfigure}[b]{0.49\textwidth}
\centering
\includegraphics[width=0.8\textwidth]{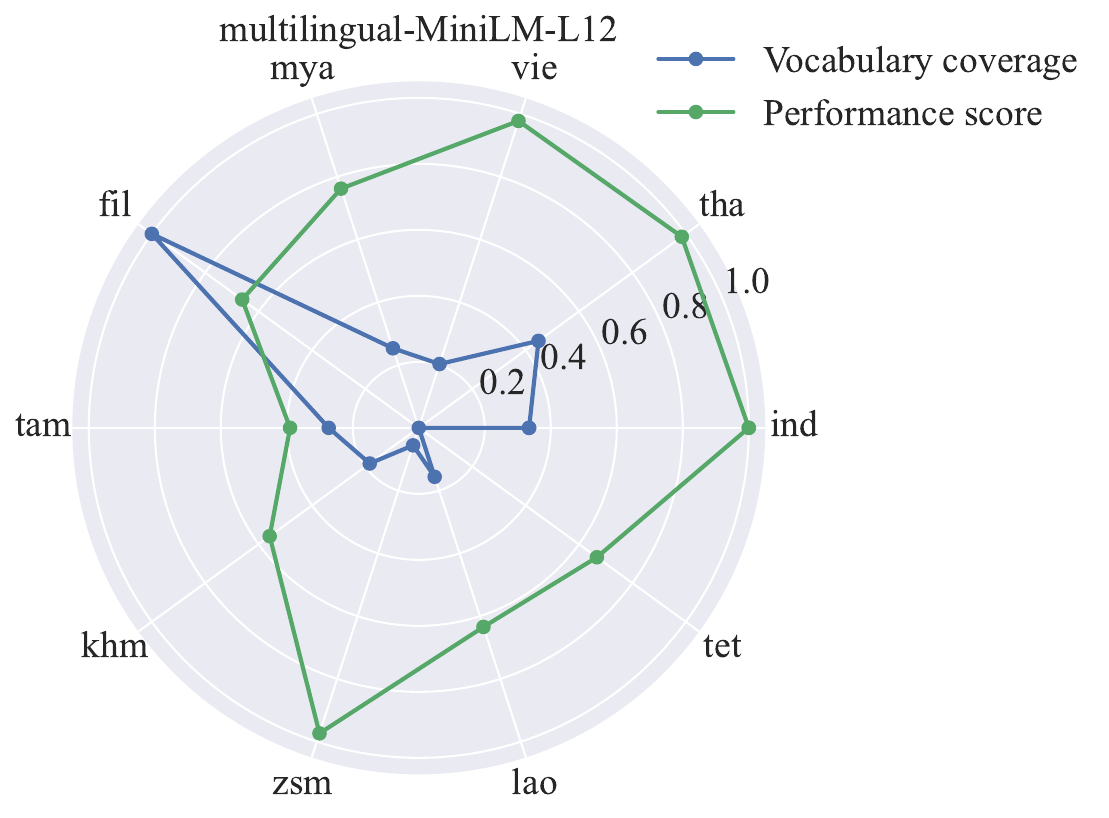}
\caption{}
\label{subfig:multilingual-MiniLM-L12_performance_vs_vocab}
\end{subfigure}
\caption{(Top) Correlation between the percentage of vocabulary token coverage and performance score for the two top-performing models, multilingual-e5-large-instruct and Qwen3-Embedding-8B.
(Bottom) Correlation between the percentage of vocabulary token coverage and performance score for the two lowest-performing models, Qwen3-Embedding-0.6B and multilingual-MiniLM-L12.
Both values are normalized to a [0, 1] scale for comparability across languages and models. 
}
\vspace{-2mm}
\label{fig:performance_vs_vocab}
\end{figure*}

\noindent
\textbf{Performance Analysis}.
In addition to studying in Appendix~\ref{appendix:translation_vs_human}, we analyze the correlation between the percentage of vocabulary token coverage and performance scores for the two top-performing and two lowest-performing embedding models, as shown in Figure~\ref{fig:performance_vs_vocab}.
The results indicate that the vocabulary size of each model does not have a direct effect on model performance in the text embedding benchmark for SEA languages.
Although some models have a larger number of tokens in their tokenizers covering SEA vocabularies, their performance in the benchmark is not significantly higher than that of models with lower vocabulary coverage.
This indicates that simply increasing vocabulary size does not necessarily lead to better performance in text embedding tasks for SEA languages.

\noindent
\textbf{Discussion}.
In contrast to previous works, we summarize that the number of tokens present in the tokenizer might not strongly correlate with the performance in a language.
There are many SEA languages with diverse scripts, and solely having a larger vocabulary for each language might not necessarily yield significant improvement.
As shown in the language performances of GritLM-7B and bge-multilingual-gemma2 (Table~\ref{tab:sea-mteb_language_results}), omitting SEA languages from the training data results in poor performance in those languages.
To achieve a promising result, we can add more SEA training datasets in the training step to improve downstream task performance rather than adding more tokens in the tokenizer.
Moreover, we report dataset statistics across Southeast Asian languages in Table~\ref{tab:token_distribution}, including the number of tokens and samples for each language, to support decision making when adding datasets for each language.

\begin{table*}[h!]
\vspace{-1mm}
\centering
\setlength\doublerulesep{4pt}
\scalebox{0.9}{
    \renewcommand{\arraystretch}{1.2}
    \begin{tabular}{lcc}
        \toprule
        \textbf{Language Name} & \textbf{Number of tokens} & \textbf{Number of samples} 
        \\
        \bottomrule
        \midrule
        Indonesian & 56,742,156  & 1,056,710 \\
        Thai & 214,798,897 & 1,337,357 \\
        Vietnamese & 33,534,467 & 910,496 \\
        Burmese & 21,080,124 & 234,913 \\
        Filipino & 13,624,124 & 248,089 \\
        Khmer & 2,0661,441 & 194,112 \\
        Malay & 15,792,761 & 365,304  \\
        Lao & 15,196,896 & 168,032 \\
        Tamil & 96,633,008 & 322,393 \\
        Tetum & 88,928,018 & 48,728 \\
        \bottomrule
    \end{tabular}
    }
    \vspace{-2mm}
    \caption{Dataset statistics across Southeast Asian languages, including the number of tokens and samples per language. 
    All statistics are computed using the tokenizer of multilingual-e5-large-instruct, the top-performing model in our evaluation.}
    \vspace{-3mm}
    \label{tab:token_distribution}
\end{table*}

\section{Models} \label{appendix:models}
To evaluate text embedding on SEA texts, we experiment on 13 open-source models across encoder and decoder models as follows:
\begin{compactitem} 
    \item \textbf{multilingual-e5-large}~\cite{wang2024multilingual}. A multilingual-e5-large model that is trained on over 100 languages using a combination of contrastive pre-training on diverse multilingual text pairs and supervised fine-tuning on high-quality labeled datasets using mined hard negatives and knowledge distillation techniques.
    \item \textbf{multilingual-e5-large-instruct}~\cite{wang2024multilingual}. The multilingual-e5-large-instruct model is similar to multilingual-e5-large, with additional fine-tuning on instructional data. 
    \item \textbf{e5-mistral-7b-instruct}~\cite{wang-etal-2024-improving-text}. The e5-mistral-7b-instruct model is a text embedding model based on Mistral-7B~\cite{jiang2023mistral7b}, fine-tuned with contrastive learning on synthetic instruction data across 93 languages. Using a two-step prompting strategy, the model learns from diverse embedding tasks and achieves strong multilingual performance with under 1,000 training steps.
    \item \textbf{multilingual-mpnet-base-v2}~\cite{reimers-2020-multilingual-sentence-bert}. The multilingual-mpnet-base-v2 model is trained on parallel data for over 50 languages via multilingual knowledge distillation using paraphrase-mpnet-base-v2~\cite{reimers-2019-sentence-bert} as a teacher model, xlm-roberta-base~\cite{DBLP:journals/corr/abs-1911-02116} as a student model, and MSE loss to align their embeddings.
    \item \textbf{LaBSE}~\cite{Feng2022labse}. The LaBSE model is trained on over 109 languages using a dual-encoder transformer architecture based on BERT~\cite{DBLP:journals/corr/abs-1810-04805}, leveraging a translation ranking loss function to produce sentence embeddings that align semantically similar sentences across languages into a shared vector space
    \item \textbf{multilingual-MiniLM-L12-v2}~\cite{reimers-2020-multilingual-sentence-bert}. The multilingual-MiniLM-L12-v2 model is trained using a similar multilingual knowledge distillation approach to multilingual-mpnet-base-v2, with paraphrase-MiniLM-L12-v2~\cite{reimers-2019-sentence-bert} as a teacher model, Multilingual-MiniLM-L12-H384~\cite{wang2020minilm} as a student model, and MSE loss to align their embeddings.
    \item \textbf{bge-m3}~\cite{bge-m3}. The BGE-M3 model is trained on over 100 languages using a combination of contrastive pre-training on diverse multilingual corpora and supervised fine-tuning with high-quality labeled and synthetic datasets, leveraging hard negative mining and a self-knowledge distillation framework that integrates dense, sparse, and multi-vector retrieval signals.
    \item \textbf{bge-multilingual-gemma2}~\cite{bge-m3}. The bge-multilingual-gemma2 model is built on Gemma-2-9b~\cite{gemma_2024} and trained on diverse multilingual data across tasks such as retrieval, classification, and clustering using embedding techniques.
    \item \textbf{GritLM-7B}~\cite{muennighoff2024generative}. The GritLM-7B model is built on the Mistral-7B~\cite{jiang2023mistral7b} architecture and trained using Generative Representational Instruction Tuning (GRIT), a unified framework combining contrastive learning for embeddings and next-token prediction for generation, with task-specific instructions and a joint loss to enable strong performance across both tasks.
    \item \textbf{Qwen3-Embedding-0.6B and Qwen3-Embedding-8B}~\cite{qwen3embedding}. The Qwen3-Embedding-0.6B and Qwen3-Embedding-8B models were trained on multiple languages using a multi-stage training pipeline that combines large-scale weakly supervised pre-training on synthetic multilingual data with supervised fine-tuning and model merging techniques to enhance robustness and generalization.
    \item \textbf{Sailor2-8B-Chat}~\cite{sailor2report}. The Sailor2-8B-Chat model, based on an expanded Qwen2.5-7B~\cite{qwen2.5}, was trained on 13 SEA languages using two-stage continual pre-training with balanced and high-quality data, followed by two-stage instruction tuning and preference tuning with length-regularized DPO~\cite{park2024disentanglinglengthqualitydirect}.
    \item \textbf{Gemma-SEA-LION-v3-9B-IT}~\cite{sea_lion_2024}. The Gemma-SEA-LION-v3-9B-IT model is fine-tuned from the Gemma2 9B~\cite{riviere2024gemma2} base model on English and multiple SEA languages (such as Indonesian, Thai, and Vietnamese), using a combination of full parameter fine-tuning, on-policy alignment, and model merging techniques.
\end{compactitem}

Moreover, we also evaluate the performance of proprietary models as follows:
\begin{compactitem}
    \item \textbf{text-embedding-3-small}. We evaluate the text-embedding-3-small~\footnote{\url{https://openai.com/index/new-embedding-models-and-api-updates}} model, which provides a highly efficient embedding model suitable for various downstream applications.
    \item \textbf{embed-multilingual-v3.0}. We evaluate the embed-multilingual-v3.0~\footnote{\url{https://cohere.com/blog/introducing-embed-v3}} model, designed for multilingual representation learning across over 100 languages.
    \item \textbf{voyage-3}. We evaluate the voyage-3~\footnote{\url{https://blog.voyageai.com/2024/09/18/voyage-3/}} model, which provides efficient, high-quality embeddings optimized for retrieval across diverse domains.
    \item \textbf{jina-embeddings-v3}. We evaluate the jina-embeddings-v3~\cite{sturua2024jinaembeddingsv3multilingualembeddingstask} model, which is designed for efficient semantic similarity and search applications, supporting various multilingual scenarios.
\end{compactitem}
All proprietary models were accessed and evaluated using their latest publicly available versions during experimentation (April 4th, 2025).
The full model links are shown in Table~\ref{tab:model_and_link}.


\begin{table*}[h]
\hspace{-4mm}
\centering
\setlength\doublerulesep{4pt}
\scalebox{0.6}{
    \renewcommand{\arraystretch}{1.2}
    \begin{tabular}{l|l}
        \toprule
        \textbf{Model} & \multicolumn{1}{c}{\textbf{Hugging Face Link}} \\
        \bottomrule
        \midrule
        multilingual-e5-large-instruct & \url{https://huggingface.co/intfloat/multilingual-e5-large-instruct} \\
        Qwen3-Embedding-8B & \url{https://huggingface.co/Qwen/Qwen3-Embedding-8B} \\
        bge-multilingual-gemma2 & \url{https://huggingface.co/BAAI/bge-multilingual-gemma2} \\
        multilingual-e5-large & \url{https://huggingface.co/intfloat/multilingual-e5-large} \\
        bge-m3 & \url{https://huggingface.co/BAAI/bge-m3} \\
        GritLM-7B & \url{https://huggingface.co/GritLM/GritLM-7B} \\
        e5-mistral-7b-instruct & \url{https://huggingface.co/intfloat/e5-mistral-7b-instruct} \\
        Qwen3-Embedding-0.6B & \url{https://huggingface.co/Qwen/Qwen3-Embedding-0.6B} \\
        multilingual-mpnet-base & \url{https://huggingface.co/sentence-transformers/paraphrase-multilingual-mpnet-base-v2} \\
        LaBSE & \url{https://huggingface.co/sentence-transformers/LaBSE} \\
        multilingual-MiniLM-L12 & \url{https://huggingface.co/sentence-transformers/paraphrase-multilingual-MiniLM-L12-v2} \\
        Gemma-SEA-LION-v3-9B-IT & \url{https://huggingface.co/aisingapore/Gemma-SEA-LION-v3-9B-IT} \\
        Sailor2-8B-Chat & \url{https://huggingface.co/sail/Sailor2-8B-Chat} \\
        \bottomrule
    \end{tabular}
    }
    \caption{Models and Hugging Face links used for the evaluation.}
    \label{tab:model_and_link}
    \vspace{-2mm}
\end{table*}

\begin{table*}[h!]
\centering
\setlength\doublerulesep{4pt}
\begin{subtable}[t]{\textwidth}
\centering
\scalebox{0.75}{
\setlength{\tabcolsep}{5pt}
\renewcommand{\arraystretch}{1.3}
\begin{tabular}{lccccccccccc}
\toprule
\textbf{Model} & \textbf{ind} & \textbf{tha} & \textbf{vie} & \textbf{mya} & \textbf{fil} & \textbf{khm} & \textbf{zsm} & \textbf{lao} & \textbf{tam} & \textbf{tet} & \textbf{Avg.} \\
\midrule
\textit{Number of datasets (→)} & (51) & (50) & (36) & (32) & (28) & (18) & (15) & (16) & (17) & (2) & (265) \\
\midrule
multilingual-e5-large-instruct (560M) & 82.06 & \textbf{83.12} & 79.54 & \textbf{78.27} & \underline{80.33} & \textbf{79.63} & \underline{88.68} & \textbf{87.07} & 76.89 & \underline{41.06} & \textbf{77.67}$_{\pm12.71}$ \\
Qwen3-Embedding-8B (8B) & 82.98 & \underline{82.85} & \textbf{80.67} & \underline{75.35} & 78.78 & 75.86 & 86.87 & 80.70 & 75.66 & 36.00 & \underline{75.57}$_{\pm13.65}$ \\
bge-m3 (568M) & 81.28 & 79.67 & 77.48 & 73.70 & 76.87 & \underline{76.56} & 88.07 & \underline{85.03} & 77.70 & 36.91 & 75.33$_{\pm13.43}$ \\
multilingual-e5-large (560M) & 81.90 & 81.72 & \textbf{80.67} & 70.56 & 79.42 & 72.63 & 82.92 & 82.81 & \underline{78.03} & 32.24 & 74.29$_{\pm14.58}$ \\
bge-multilingual-gemma2 (9B) & \underline{83.28} & 82.08 & \underline{80.29} & 70.43 & \textbf{80.67} & 74.03 & 85.18 & 66.50 & \textbf{80.99} & 30.91 & 73.44$_{\pm15.27}$ \\
LaBSE (471M) & 75.72 & 72.59 & 74.82 & 73.98 & 78.27 & 74.66 & \textbf{89.21} & 83.41 & 77.05 & \textbf{43.46} & 74.42$_{\pm11.34}$ \\
multilingual-mpnet-base (278M) & 77.09 & 76.06 & 74.01 & 60.52 & 51.01 & 62.34 & 77.23 & 65.09 & 63.15 & 34.54 & 64.10$_{\pm12.83}$ \\
e5-mistral-7b-instruct (7B) & 82.40 & 76.50 & 77.32 & 46.95 & 79.17 & 54.68 & 81.45 & 23.12 & 66.61 & 37.21 & 62.54$_{\pm20.00}$ \\
GritLM-7B (7B) & \textbf{83.32} & 74.64 & 78.80 & 42.56 & 78.48 & 50.20 & 80.76 & 24.36 & 60.09 & 40.72 & 61.39$_{\pm19.77}$ \\
Qwen3-Embedding-0.6B (595M) & 78.83 & 77.16 & 76.70 & 47.62 & 62.83 & 39.19 & 71.48 & 24.31 & 60.67 & 30.69 & 56.95$_{\pm19.22}$ \\
multilingual-MiniLM-L12 (118M) & 73.54 & 72.49 & 71.26 & 53.68 & 46.20 & 35.01 & 70.63 & 42.72 & 27.57 & 30.54 & 52.36$_{\pm17.52}$ \\
Gemma-SEA-LION-v3-9B-IT (9B) & 51.93 & 42.10 & 52.09 & 28.02 & 53.69 & 35.63 & 52.25 & 16.07 & 27.58 & 0.08 & 35.94$_{\pm17.17}$ \\
Sailor2-8B-Chat (8B) & 51.30 & 36.18 & 42.23 & 27.43 & 45.19 & 22.91 & 29.08 & 11.84 & 27.22 & 1.45 & 29.48$_{\pm14.42}$ \\
\midrule
\emph{Proprietary models} \\ \midrule
embed-multilingual-v3.0 & \textbf{83.03} & \textbf{82.94} & \textbf{80.67} & \textbf{76.94} & \textbf{80.16} & \textbf{78.02} & \textbf{86.66} & \textbf{86.64} & \textbf{78.86} & \textbf{38.08} & \textbf{77.20}$_{\pm13.42}$ \\
jina-embeddings-v3 & 80.37 & \underline{80.35} & \underline{77.25} & \underline{75.61} & \underline{74.94} & \underline{74.59} & \underline{81.96} & \underline{79.56} & \underline{75.94} & 33.89 & \underline{73.45}$_{\pm13.41}$ \\
voyage-3 & 78.28 & 71.60 & 75.25 & 46.42 & 72.18 & 29.32 & 73.03 & 18.49 & 67.37 & 25.62 & 55.76$_{\pm22.19}$ \\
text-embedding-3-small & \underline{82.04} & 55.86 & 71.64 & 30.68 & 68.52 & 24.25 & 71.76 & 18.36 & 34.01 & \underline{35.25} & 49.24$_{\pm22.02}$ \\
\bottomrule
\end{tabular}
}
\caption{Human-crafted datasets}
\end{subtable}

\vspace{6mm}
\begin{subtable}[t]{\textwidth}
\centering
\scalebox{0.75}{
\setlength{\tabcolsep}{5pt}
\renewcommand{\arraystretch}{1.3}
\begin{tabular}{lccccccccccc}
\toprule
\textbf{Model} & \textbf{ind} & \textbf{tha} & \textbf{vie} & \textbf{mya} & \textbf{fil} & \textbf{khm} & \textbf{zsm} & \textbf{lao} & \textbf{tam} & \textbf{tet} & \textbf{Avg.} \\
\midrule
\textit{Number of datasets (→)} & (19) & (5) & (5) & (3) & (3) & (4) & (4) & (3) & (1) & (2) & (49) \\
\midrule
multilingual-e5-large-instruct (560M) & \underline{72.04} & 61.03 & 66.92 & \textbf{79.47} & 68.54 & 71.38 & 69.31 & 67.23 & \underline{80.45} & 97.73 & \underline{73.41}$_{\pm9.79}$ \\
Qwen3-Embedding-8B (8B) & 69.95 & \textbf{67.83} & 66.91 & 70.19 & \textbf{71.24} & 73.66 & 65.62 & 64.85 & \textbf{80.86} & \underline{98.89} & 73.00$_{\pm9.68}$ \\
bge-m3 (568M) & 69.14 & 56.74 & 64.57 & 66.90 & 65.62 & \underline{74.74} & 61.77 & \underline{67.47} & 74.39 & 94.15 & 69.55$_{\pm9.65}$ \\
multilingual-e5-large (560M) & 68.58 & 61.60 & 66.43 & 67.30 & 64.61 & 69.77 & 69.52 & 64.45 & 74.34 & 94.86 & 70.15$_{\pm8.89}$ \\
bge-multilingual-gemma2 (9B) & 71.17 & \underline{65.64} & \textbf{67.80} & 65.55 & \underline{69.77} & \textbf{76.00} & \textbf{76.63} & 62.19 & 80.40 & \textbf{99.18} & \textbf{73.43}$_{\pm10.14}$ \\
LaBSE (471M) & 68.42 & 46.30 & 56.59 & 69.90 & 65.05 & 71.38 & 59.14 & 60.84 & 68.88 & 94.85 & 66.14$_{\pm12.01}$ \\
multilingual-mpnet-base (278M) & 67.93 & 52.53 & 62.97 & 68.37 & 61.37 & 73.89 & 68.93 & \textbf{68.51} & 66.07 & 67.01 & 65.76$_{\pm5.47}$ \\
e5-mistral-7b-instruct (7B) & 70.13 & 57.49 & 61.31 & 69.09 & 68.11 & 64.61 & 68.93 & 53.93 & 68.74 & 96.26 & 67.86$_{\pm10.82}$ \\
GritLM-7B (7B) & \textbf{72.24} & 54.86 & \underline{67.10} & \underline{71.61} & 68.20 & 63.31 & \underline{69.58} & 60.49 & 66.04 & 98.62 & 69.21$_{\pm11.01}$ \\
Qwen3-Embedding-0.6B (595M) & 66.02 & 62.77 & 63.78 & 64.62 & 65.68 & 66.20 & 62.14 & 58.99 & 67.51 & 96.07 & 67.38$_{\pm9.84}$ \\
multilingual-MiniLM-L12 (118M) & 65.86 & 49.79 & 60.10 & 62.95 & 56.71 & 61.99 & 65.64 & 59.30 & 33.12 & 64.84 & 58.03$_{\pm9.50}$ \\
Gemma-SEA-LION-v3-9B-IT (9B) & 44.34 & 37.41 & 50.56 & 60.52 & 58.38 & 57.09 & 37.94 & 53.72 & 56.71 & 50.04 & 50.67$_{\pm7.89}$ \\
Sailor2-8B-Chat (8B) & 44.86 & 33.92 & 48.10 & 59.03 & 55.23 & 56.18 & 36.99 & 52.86 & 51.61 & 50.08 & 48.89$_{\pm7.77}$ \\
\midrule
\emph{Proprietary models} \\ \midrule
embed-multilingual-v3.0 & \textbf{69.76} & \underline{61.41} & \underline{66.40} & \underline{67.52} & \textbf{68.06} & \underline{72.48} & \textbf{66.53} & \underline{65.73} & \underline{79.09} & 95.43 & \underline{71.24}$_{\pm9.20}$ \\
jina-embeddings-v3 & \underline{68.40} & \textbf{61.53} & \textbf{67.83} & \textbf{69.73} & \underline{67.76} & \textbf{75.37} & \underline{62.74} & \textbf{69.11} & \textbf{79.70} & \underline{96.33} & \textbf{71.85}$_{\pm9.59}$ \\
voyage-3 & 67.32 & 51.57 & 62.38 & 67.12 & 64.41 & 60.69 & 54.49 & 55.08 & 65.71 & \textbf{97.34} & 64.61$_{\pm12.12}$ \\
text-embedding-3-small & 67.53 & 49.03 & 58.65 & 55.30 & 63.93 & 56.71 & 62.36 & 53.88 & 58.56 & 94.92 & 62.09$_{\pm12.03}$ \\
\bottomrule
\end{tabular}
}
\caption{Machine-generated datasets}
\end{subtable}

\vspace{-1mm}
\caption{Language-model performance view, where each cell reports scores averaged over all evaluated task types, separated into human-crafted datasets (Top) and machine-generated datasets (Bottom).}
\vspace{3mm}
\label{tab:sea-bed-language-results}
\end{table*}

\begin{table*}[h!]
\hspace*{-3.5mm}
\centering
\setlength\doublerulesep{4pt}
\begin{subtable}[t]{\textwidth}
\centering
\scalebox{0.68}{
\setlength{\tabcolsep}{5pt}
\renewcommand{\arraystretch}{1.2}
\begin{tabular}{lc|cccccccccc}
\toprule
\textbf{Model} & \textbf{Dim.} &
\textbf{Clf} & \textbf{M. Clf} & \textbf{Pr. Clf} & \textbf{STS} &
\textbf{Clust} & \textbf{Btxt} & \textbf{Rtrvl} & \textbf{In. Rtrvl} & \textbf{Rrnk} &
\textbf{Avg.} \\
\bottomrule
\textit{Number of datasets (→)} &
& (64) & (9) & (7) & (10) & (10) & (20) & (18) & (1) & (1) & (140) \\
\midrule
multilingual-e5-large-instruct (560M) & 1024
& 80.33 & 86.61 & 69.10 & \underline{73.08}
& \textbf{53.18} & \textbf{87.62} & 80.41 & \underline{96.38} & 77.24
& \textbf{78.22}$_{\pm11.72}$ \\

Qwen3-Embedding-8B (8B) & 4096
& \textbf{81.48} & \underline{89.35} & 65.45 & \textbf{73.38}
& \underline{43.91} & 84.67 & \textbf{83.89} & 96.18 & \underline{78.51}
& \underline{77.42}$_{\pm14.49}$ \\

bge-m3 (568M) & 4096
& 78.59 & 88.78 & 71.88 & 71.49
& 33.73 & 86.34 & 77.92 & 87.73 & 75.98
& 74.72$_{\pm15.73}$ \\

multilingual-e5-large (560M) & 1024
& 80.97 & 87.87 & 67.63 & 68.07
& 37.03 & 84.51 & 81.66 & 96.01 & \textbf{79.00}
& 75.86$_{\pm16.09}$ \\

bge-multilingual-gemma2 (9B) & 3584
& \underline{81.19} & \textbf{89.44} & \textbf{78.53} & 69.57
& 41.98 & 81.86 & \underline{82.81} & \textbf{96.89} & 69.04
& 76.81$_{\pm14.80}$ \\

LaBSE (471M) & 768
& 77.53 & 85.13 & 63.45 & 67.92
& 32.53 & \underline{86.61} & 56.82 & 79.92 & 61.23
& 67.90$_{\pm16.06}$ \\

multilingual-mpnet-base (278M) & 768
& 76.18 & 85.56 & \underline{75.02} & 67.84
& 33.87 & 67.67 & 61.05 & 86.80 & 64.01
& 68.67$_{\pm14.92}$ \\

e5-mistral-7b-instruct (7B) & 4096
& 79.03 & 86.96 & 65.99 & 60.77
& 40.91 & 64.06 & 75.51 & 94.31 & 75.33
& 71.43$_{\pm14.85}$ \\

GritLM-7B (7B) & 4096
& 79.80 & 87.16 & 65.82 & 62.27
& 36.67 & 62.15 & 68.64 & 93.50 & 73.37
& 69.93$_{\pm15.66}$ \\

Qwen3-Embedding-0.6B (595M) & 1024
& 77.02 & 86.83 & 62.33 & 64.14
& 34.79 & 55.19 & 78.23 & 94.30 & 75.03
& 69.76$_{\pm16.89}$ \\

multilingual-MiniLM-L12 (118M) & 768
& 72.85 & 83.02 & 69.86 & 64.23
& 23.51 & 52.14 & 55.10 & 83.58 & 62.27
& 62.95$_{\pm17.35}$ \\

Gemma-SEA-LION-v3-9B-IT (9B) & 3584
& 78.67 & 88.34 & 57.89 & 32.62
& 28.74 & 16.05 & 23.74 & 31.16 & 65.49
& 46.97$_{\pm24.64}$ \\

Sailor2-8B-Chat (8B) & 3584
& 79.44 & 88.80 & 56.63 & 32.32
& 26.28 & 4.25 & 10.94 & 10.57 & 47.05
& 39.59$_{\pm28.86}$ \\

\midrule
\emph{Proprietary models} \\ \midrule
embed-multilingual-v3.0 & 1024
& \textbf{81.59} & \textbf{88.87} & \textbf{68.29} & \textbf{71.24}
& \underline{40.69} & \textbf{88.34} & \textbf{81.36} & \textbf{96.87} & \textbf{77.77}
& \textbf{77.22}$_{\pm15.39}$ \\

jina-embeddings-v3 & 1024
& \underline{80.25} & \underline{87.72} & \underline{65.98} & \underline{69.56}
& \textbf{44.38} & \underline{81.83} & \underline{78.94} & \underline{96.51} & 72.49
& \underline{75.30}$_{\pm14.02}$ \\

voyage-3 & 1024
& 78.26 & 87.36 & 62.59 & 61.50
& 36.50 & 54.22 & 66.19 & 87.29 & \underline{74.62}
& 67.61$_{\pm15.46}$ \\

text-embedding-3-small & 1536
& 75.76 & 86.53 & 61.95 & 47.54
& 30.24 & 40.82 & 67.57 & 83.60 & 71.25
& 62.81$_{\pm18.36}$ \\
\bottomrule
\end{tabular}
}
\caption{Human-crafted datasets}
\label{tab:sea-bed-task-human}
\end{subtable}

\vspace{2mm}
\begin{subtable}[h]{\textwidth}
\centering
\scalebox{0.68}{
\setlength{\tabcolsep}{5pt}
\renewcommand{\arraystretch}{1.2}
\begin{tabular}{lc|cccccccccc}
\toprule
\textbf{Model} & \textbf{Dim.} &
\textbf{Clf} & \textbf{M. Clf} & \textbf{Pr. Clf} & \textbf{STS} &
\textbf{Clust} & \textbf{Btxt} & \textbf{Rtrvl} & \textbf{In. Rtrvl} & \textbf{Rrnk} &
\textbf{Avg.} \\
\bottomrule
\textit{Number of datasets (→)} &
& (9) & (2) & (6) & (1) & (0) & (6) & (2) & (3) & (0) & (29) \\
\midrule
multilingual-e5-large-instruct (560M) & 1024
& 65.76 & 93.34 & 64.24 & \textbf{80.61}
& - & \textbf{92.13} & 43.04 & 60.00 & - 
& 71.30$_{\pm16.96}$ \\

Qwen3-Embedding-8B (8B) & 4096
& 65.45 & 96.05 & 60.38 & \underline{79.19}
& - & 86.67 & \textbf{61.99} & \underline{62.35} & -
& \underline{73.15}$_{\pm13.13}$ \\

bge-m3 (568M) & 4096
& 64.12 & 94.87 & 65.76 & 76.84
& - & 83.29 & 27.70 & 48.76 & -
& 65.91$_{\pm20.76}$ \\

multilingual-e5-large (560M) & 1024
& 65.79 & 93.78 & 63.00 & 72.67
& - & 84.57 & 42.48 & 56.07 & -
& 68.34$_{\pm15.96}$ \\

bge-multilingual-gemma2 (9B) & 3584
& 64.19 & \textbf{97.45} & \textbf{70.37} & 78.44
& - & 84.81 & \underline{56.81} & \textbf{63.07} & -
& \textbf{73.59}$_{\pm13.15}$ \\

LaBSE (471M) & 768
& 64.52 & 93.48 & 60.60 & 69.13
& - & \underline{90.89} & 21.12 & 26.34 & -
& 60.87$_{\pm26.24}$ \\

multilingual-mpnet-base (278M) & 768
& 62.87 & 94.99 & \underline{67.01} & 74.76
& - & 75.91 & 29.14 & 40.98 & -
& 63.67$_{\pm20.61}$ \\

e5-mistral-7b-instruct (7B) & 4096
& \underline{65.81} & 94.43 & 62.16 & 68.97
& - & 87.72 & 45.87 & 41.18 & -
& 66.59$_{\pm18.22}$ \\

GritLM-7B (7B) & 4096
& \textbf{66.83} & 95.98 & 62.50 & 69.54
& - & 90.38 & 37.93 & 58.96 & -
& 68.87$_{\pm18.12}$ \\

Qwen3-Embedding-0.6B (595M) & 1024
& 62.89 & 94.34 & 58.17 & 68.95
& - & 80.57 & 55.39 & 56.30 & -
& 68.09$_{\pm13.48}$ \\

multilingual-MiniLM-L12 (118M) & 768
& 59.81 & 93.24 & 62.21 & 65.30
& - & 73.23 & 24.94 & 37.01 & -
& 59.39$_{\pm20.94}$ \\

Gemma-SEA-LION-v3-9B-IT (9B) & 3584
& 63.11 & \underline{97.12} & 56.86 & 51.31
& - & 2.03 & 4.12 & 4.31 & -
& 39.84$_{\pm34.25}$ \\

Sailor2-8B-Chat (8B) & 3584
& 62.72 & 96.58 & 56.18 & 47.11
& - & 5.28 & 1.18 & 0.86 & -
& 38.56$_{\pm34.35}$ \\

\midrule
\emph{Proprietary models} \\ \midrule
embed-multilingual-v3.0 & 1024
& \textbf{64.52} & \underline{94.97} & \textbf{63.15} & \underline{76.86}
& - & \textbf{87.86} & \underline{44.68} & \underline{55.49} & -
& \underline{69.65}$_{\pm16.55}$ \\

jina-embeddings-v3 & 1024
& \underline{64.41} & 94.61 & \underline{60.96} & \textbf{80.38}
& - & 82.28 & \textbf{48.38} & \textbf{59.98} & -
& \textbf{70.14}$_{\pm14.86}$ \\

voyage-3 & 1024
& 64.13 & 94.69 & 57.55 & 62.92
& - & 80.80 & 28.42 & 53.27 & -
& 63.11$_{\pm19.43}$ \\

text-embedding-3-small & 1536
& 59.77 & \textbf{95.62} & 58.51 & 61.84
& - & \underline{84.55} & 39.99 & 42.63 & -
& 63.27$_{\pm18.91}$ \\
\bottomrule
\end{tabular}
}
\caption{Machine-generated datasets}
\label{tab:sea-bed-task-machine}
\end{subtable}

\vspace{-1mm}
\caption{Task-model performance view, where each cell reports scores averaged over all evaluated languages, separated into human-crafted datasets (Top) and machine-generated datasets (Bottom). “-” indicates that no dataset is available for the corresponding task.}
\vspace{-1mm}
\label{tab:sea-bed-task-results}
\end{table*}

\begin{table*}[h!]
\centering
\setlength\doublerulesep{4pt}

\begin{subtable}[t]{\textwidth}
\centering
\scalebox{0.7}{
\setlength{\tabcolsep}{5pt}
\renewcommand{\arraystretch}{1.2}
\begin{tabular}{l|cccccccccc}
\toprule
\textbf{Model} & \textbf{Btxt} & \textbf{Clf} & \textbf{Clust} & \textbf{In. Rtrvl} & \textbf{M. Clf} & \textbf{Pr. Clf} & \textbf{Rtrvl} & \textbf{Rrnk} & \textbf{STS} \\
\midrule
Indonesian & 75.98 & 81.00 & 43.55 & - & 92.49 & 76.28 & 72.71 & 69.28 & 46.43 \\
Thai & 70.99 & 75.67 & 38.49 & 82.98 & 77.00 & 70.31 & 74.69 & 71.86 & 68.32 \\
Vietnamese & 76.36 & 80.62 & 38.56 & - & 96.49 & 71.86 & 56.80 & - & - \\
Burmese & 48.83 & 85.10 & 28.09 & - & 69.71 & 67.10 & 55.44 & - & 61.19 \\
Filipino & 69.56 & 75.60 & 52.69 & - & - & 50.63 & - & - & - \\
Khmer & 54.21 & 74.39 & 23.83 & - & - & - & - & - & - \\
Malay & 75.98 & 77.21 & - & - & - & - & - & - & - \\
Lao & 51.79 & 75.91 & 18.23 & - & - & - & - & - & - \\
Tamil & 61.23 & 84.35 & 38.75 & - & - & 67.14 & 46.27 & - & 37.21 \\
Tetum & 31.09 & - & - & - & - & - & - & - & - \\
\bottomrule
\end{tabular}
}
\caption{Human-crafted datasets}
\end{subtable}

\vspace{2mm}
\begin{subtable}[t]{\textwidth}
\centering
\scalebox{0.7}{
\setlength{\tabcolsep}{5pt}
\renewcommand{\arraystretch}{1.2}
\begin{tabular}{l|cccccccccc}
\toprule
\textbf{Model} & \textbf{Btxt} & \textbf{Clf} & \textbf{Clust} & \textbf{In. Rtrvl} & \textbf{M. Clf} & \textbf{Pr. Clf} & \textbf{Rtrvl} & \textbf{Rrnk} & \textbf{STS} \\
\midrule
Indonesian & 74.84 & 67.61 & - & 34.82 & - & 59.58 & - & - & 75.65 \\
Thai & - & 59.11 & - & 57.88 & - & - & 25.87 & - & 71.99 \\
Vietnamese & - & 67.96 & - & 42.75 & - & 55.49 & - & - & 77.15 \\
Burmese & - & 66.15 & - & - & - & - & - & - & 68.03 \\
Filipino & - & 36.12 & - & - & 90.06 & - & - & - & 68.41 \\
Khmer & - & 39.36 & - & - & 100.00 & 66.41 & - & - & 63.74 \\
Malay & - & 55.84 & - & - & - & 71.50 & 46.27 & - & 75.39 \\
Lao & - & 59.81 & - & - & - & 64.36 & - & - & 59.11 \\
Tamil & - & - & - & - & - & - & - & - & 67.78 \\
Tetum & 75.08 & 99.81 & - & - & - & - & - & - & - \\
\bottomrule
\end{tabular}
}
\caption{Machine-generated datasets}
\end{subtable}

\vspace{-1mm}
\caption{Language-task performance view, where each cell reports scores averaged over all evaluated models, separated into human-crafted datasets (Top) and machine-generated datasets (Bottom). “-” indicates that no dataset is available for the corresponding language-task combination.}
\vspace{-1mm}
\label{tab:sea-bed-language-task-results}
\end{table*}

\section{Example of Our Evaluation Tool}
\begin{figure*}[h!]
\centering
\centering
\includegraphics[width=1.0\textwidth]{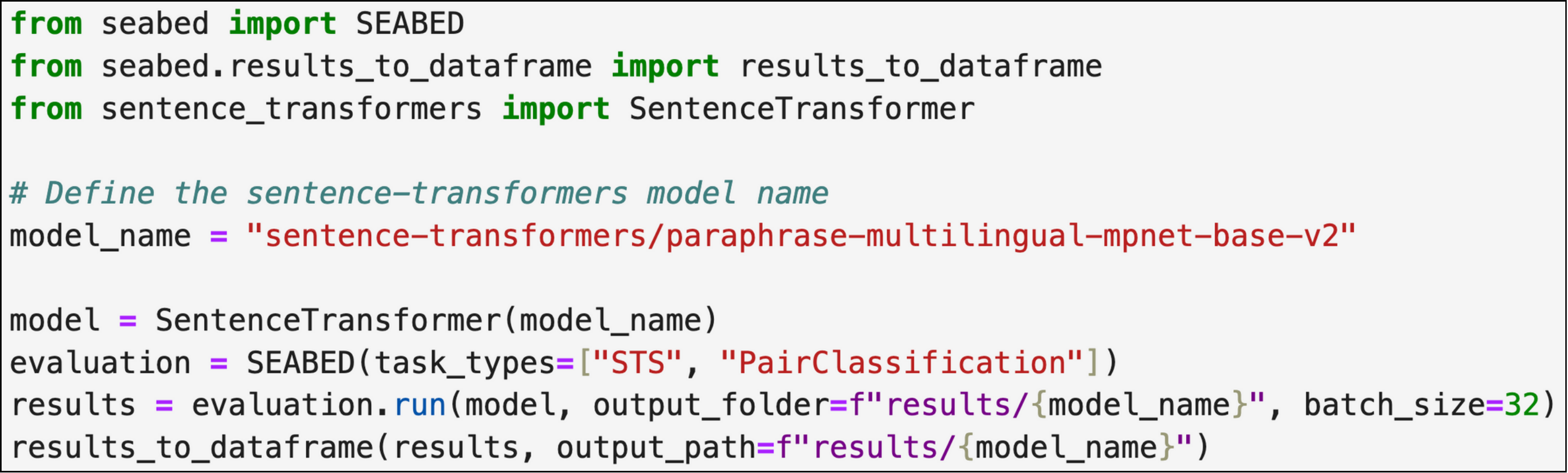}
\caption{Example usage of the SEA-BED evaluation framework for Semantic Textual Similarity (STS) and Pair Classification tasks.}
\label{fig:code_example}
\end{figure*}

Similar to the previous text embedding benchmarks~\cite{Muennighoff2023mteb,enevoldsen2025mmteb}, the evaluation tool of SEA-BED can be simply run using Python as shown in Figure~\ref{fig:code_example}.
We will release all the evaluation tools, codes, results, and datasets in the final version of our paper. 

\section{Task Examples}
\label{appendix:examples}
\begin{figure*}[h]
\centering
\vspace{-1mm}
\includegraphics[width=1.0\textwidth]{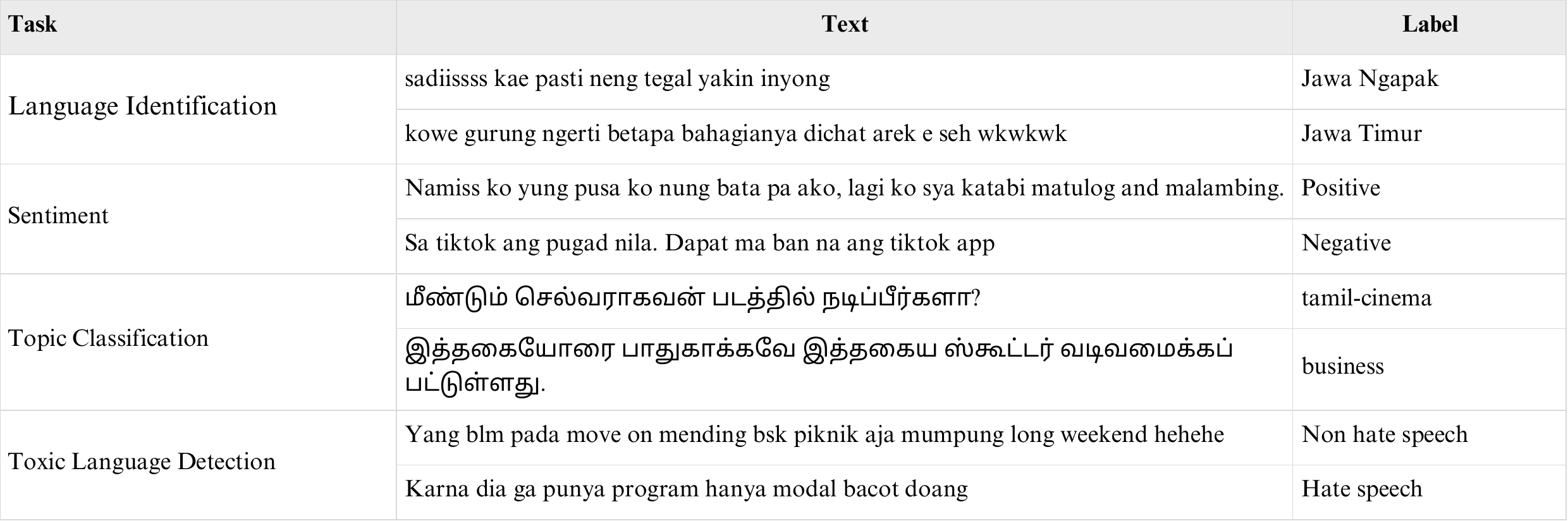}
\caption{Classification examples.}
\label{fig:classification_examples}
\end{figure*}

\begin{figure*}[h]
\vspace{-1mm}
\centering
\includegraphics[width=1.0\textwidth]{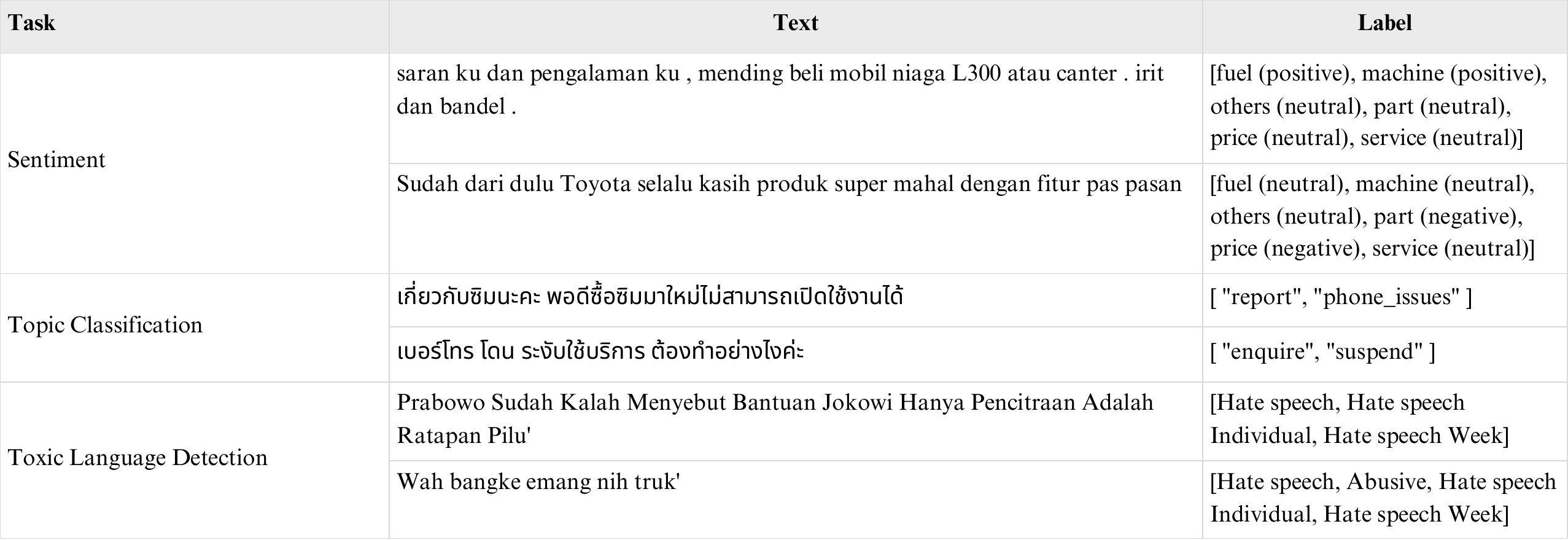}
\caption{Multi-label Classification examples.}
\label{fig:multi-label_classification_examples}
\end{figure*}

\begin{figure*}[h]
\vspace{-1mm}
\centering
\includegraphics[width=1.0\textwidth]{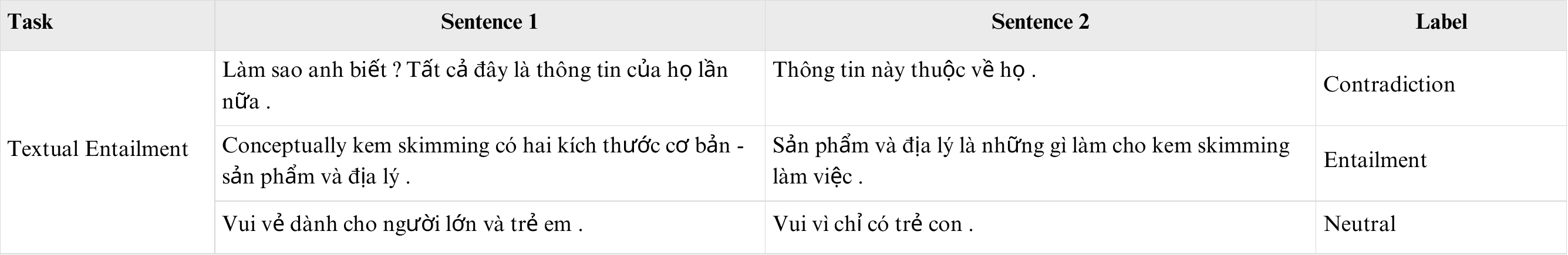}
\caption{Pair Classification examples.}
\label{fig:pair_classification_examples}
\end{figure*}

\begin{figure*}[h]
\vspace{-1mm}
\centering
\includegraphics[width=1.0\textwidth]{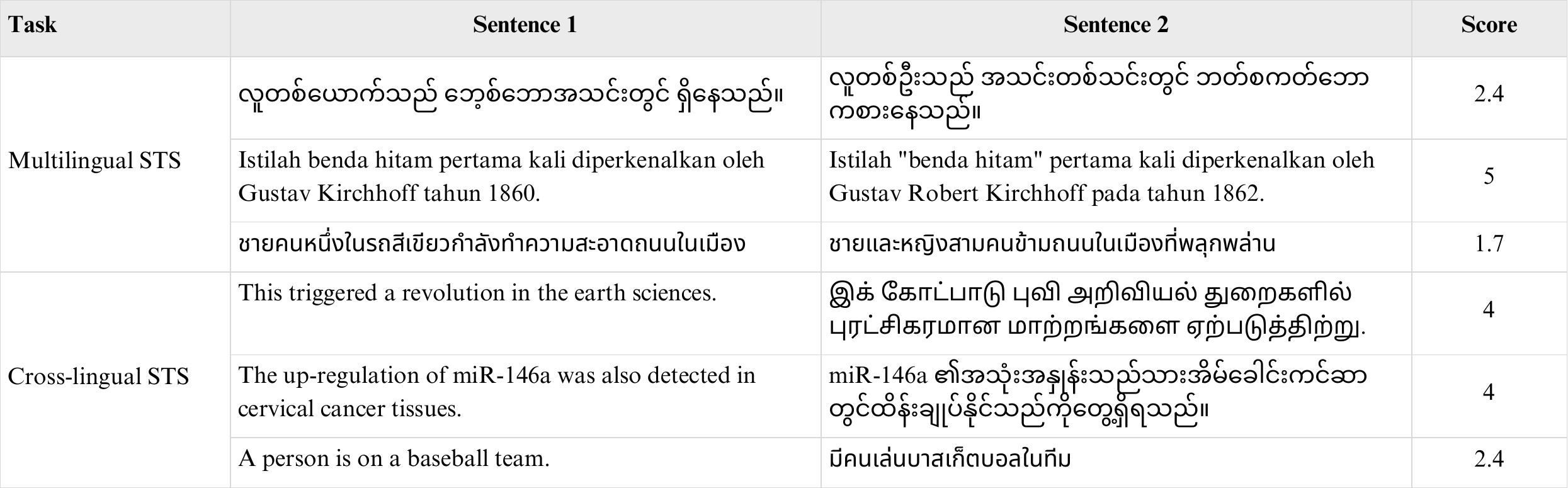}
\caption{STS examples.}
\label{fig:sts_examples}
\end{figure*}

\begin{figure*}[h]
\vspace{-1mm}
\centering
\includegraphics[width=1.0\textwidth]{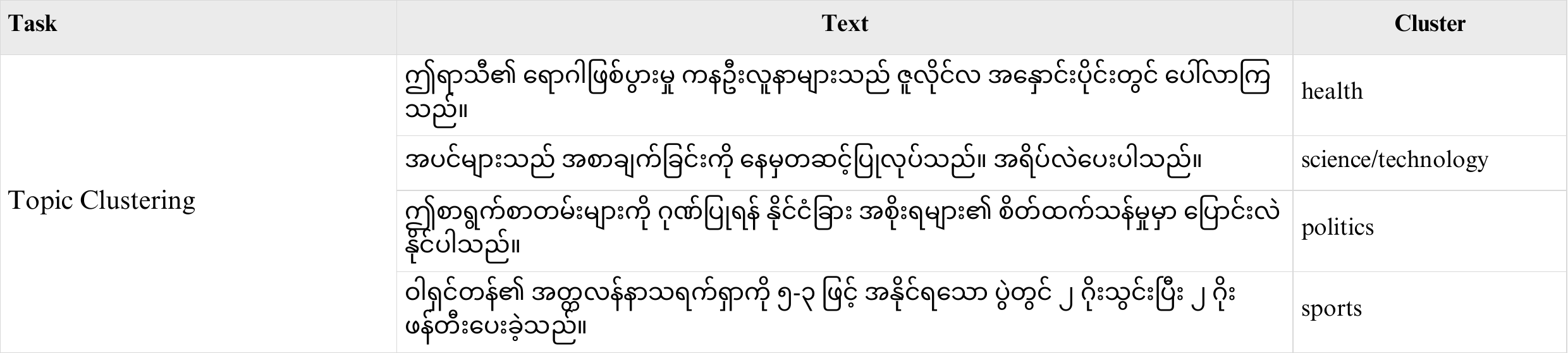}
\caption{Clustering examples.}
\label{fig:clustering_examples}
\end{figure*}

\begin{figure*}[h]
\vspace{-1mm}
\centering
\includegraphics[width=1.0\textwidth]{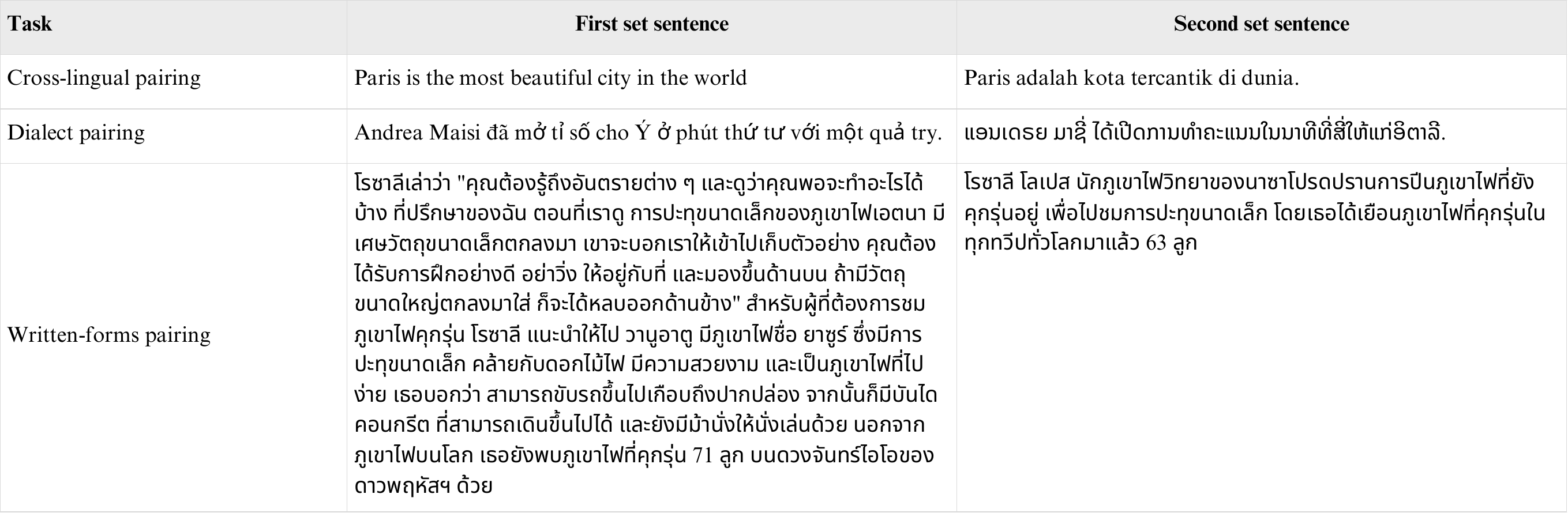}
\caption{Bitext mining examples.}
\label{fig:bitext_mining_examples}
\end{figure*}

\begin{figure*}[h]
\vspace{-4mm}
\centering
\includegraphics[width=1.0\textwidth]{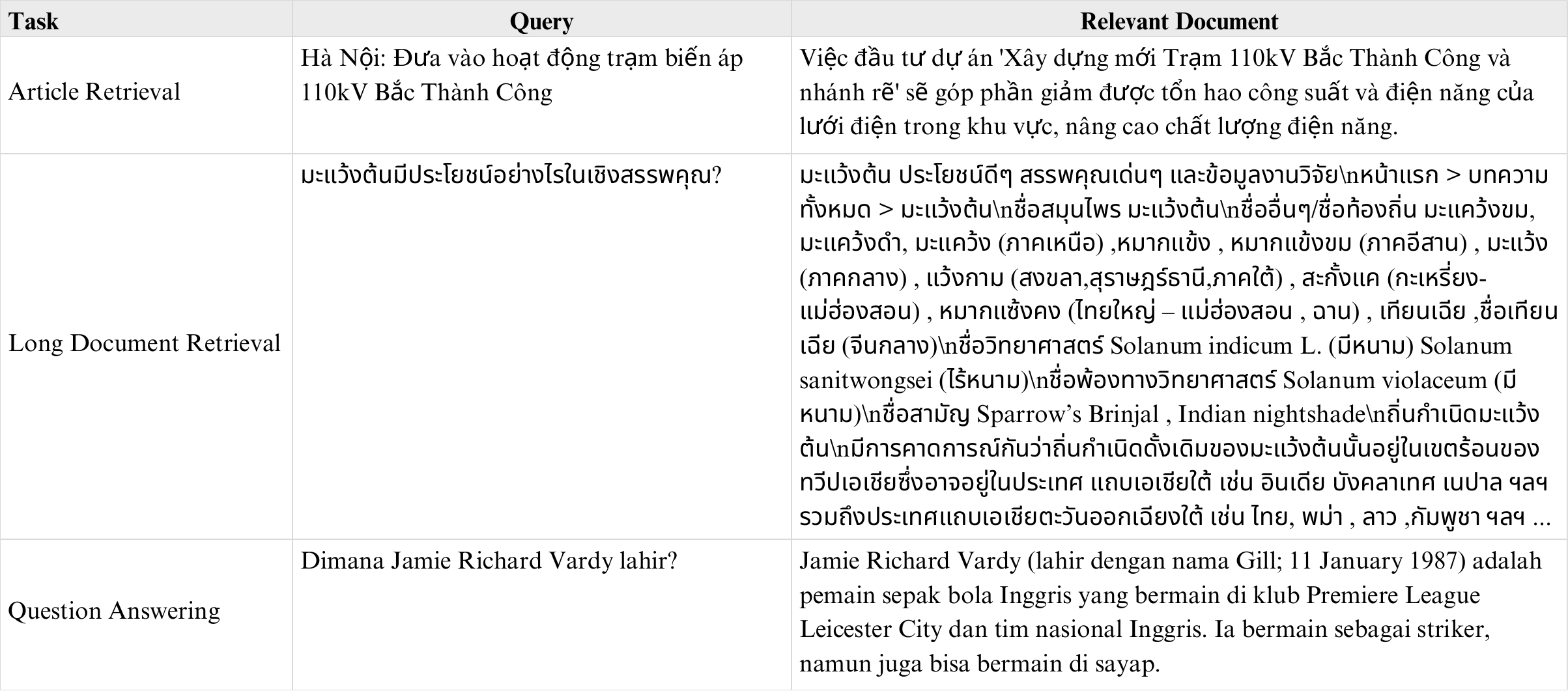}
\caption{Retrieval examples.}
\label{fig:retrieval_examples}
\end{figure*}

\begin{figure*}[h]
\vspace{-1mm}
\centering
\includegraphics[width=1.0\textwidth]{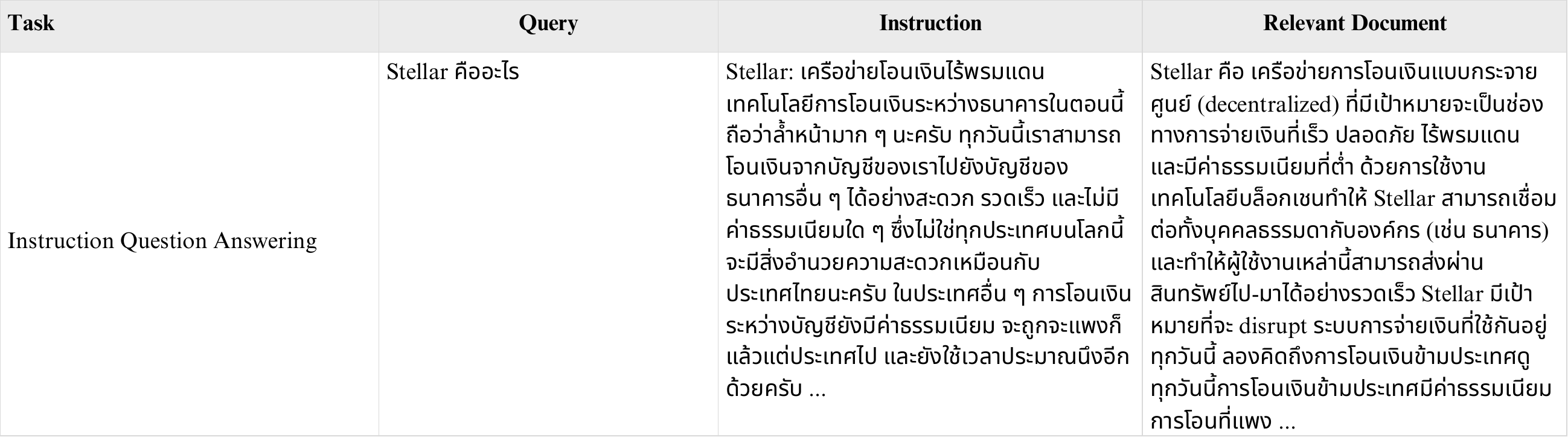}
\caption{Instruction Retrieval examples.}
\label{fig:instruction_retrieval_examples}
\end{figure*}

\begin{figure*}[h]
\vspace{-4mm}
\centering
\includegraphics[width=1.0\textwidth]{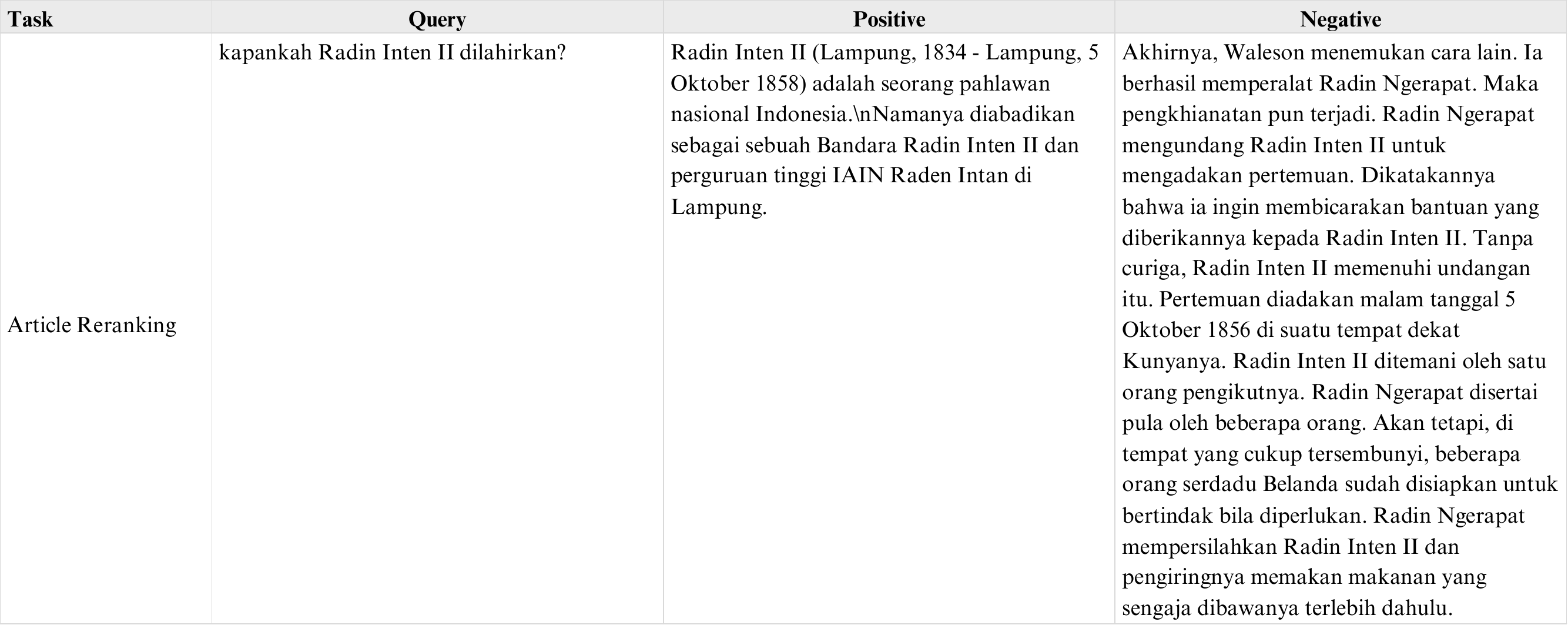}
\caption{Reranking examples.}
\label{fig:reranking_examples}
\end{figure*}

Figures~\ref{fig:classification_examples} to \ref{fig:reranking_examples} provide examples for each task covered in SEA-BED benchmark.

\section{Data Links} \label{appendix:data_links}
The complete dataset information, such as citations, languages, domains, annotation creators, and licenses, are shown in Tables~\ref{tab:datasets_in_sea-mteb_part1} and \ref{tab:datasets_in_sea-mteb_part2}.

\begin{table*}[h!]
\hspace{-4mm}
\centering
\setlength\doublerulesep{4pt}
\scalebox{0.4}{
    \renewcommand{\arraystretch}{1.2}
    \begin{tabular}{lllllll}
        \toprule
        \textbf{Type} & \textbf{Name} & \textbf{Languages} & \textbf{Domains} & \textbf{Sample creation} & \textbf{Annotations creators} & \textbf{License}
        \\
        \bottomrule
        Classification & ABUSIVE~\cite{IBROHIM2018222} & ['ind'] & ['Social', 'Written'] & found & human-annotated & CC BY-SA 4.0 \\
        & AbusiveNewsComment~\cite{9034620} & ['ind'] & ['Social', 'Web', 'News', ...] & found & human-annotated & CC BY-SA 4.0 \\
        & BookmebusReviews & ['khm'] & ['Reviews', 'Written'] & found & human-annotated &  \\
        & Clickbait\textsuperscript{*}~\cite{william2020clickid} & ['ind'] & ['News', 'Written'] & found & expert-annotated &  \\
        & CodeMixed~\cite{Tho_2021} & ['ind'] & ['Social', 'Web'] & found & manual curation & CC BY 3.0 \\
        & CyberbullyingLGBT & ['tha'] & ['Social', 'Written'] & found & derived &  \\
        & Depression~\cite{hamalainen-etal-2021-detecting} & ['tha'] & ['Social', 'Web', 'News', ...] & found & human-annotated & CC BY-NC-ND 4.0 \\
        & EMoTES3K~\cite{catapang-visperas-2023-emotion} & ['fil'] & ['Morality', 'Written'] & found & human-annotated & Apache license 2.0 \\
        & Emoji & ['tha'] & ['Social', 'Written'] & found & human-annotated & GPL-3.0 \\
        & EmoT~\cite{saputri2018emotion} & ['ind'] & ['Social', 'Written'] & found & human-annotated & MIT \\
        & EmotionOpinion~\cite{RICCOSAN2022108465} & ['ind'] & ['Social', 'Written'] & found & human-annotated & CC BY-SA 4.0 \\
        & EmotCMT~\cite{Yulianti2021NormalisationOI} & ['ind'] & ['Social', 'Written'] & found & derived & MIT \\
        & Fakenews~\cite{cruz2020localization} & ['fil'] & ['News', 'Written'] & found & human-annotated &  \\
        & GeneralAmy~\cite{phatthiyaphaibun-etal-2023-pythainlp} & ['tha'] & ['Social', 'Written'] & found & human-annotated & CC BY 3.0 \\
        & GeneratedReviewsENTH~\cite{DBLP:journals/lre/LowphansirikulP22} & ['tha'] & ['conversation', 'Web', 'Written', ...] & found & human-annotated & CC BY-SA 4.0 \\
        & GKLMIPSentiment~\cite{jiang2021myanmar} & ['mya'] & ['Social', 'Web', ''Written] & found & derived &  \\
        & GooglePlayReview & ['ind'] & ['Reviews', 'Written'] & found & human-annotated & CC BY 4.0 \\
        & HateSpeech~\cite{inproceedings} & ['ind'] & ['Social', 'Written'] & found & human-annotated &  \\
        & HateSpeech\textsuperscript{*} & ['fil'] & ['Social', 'Written'] & found & human-annotated & Apache license 2.0 \\
        & HoaxNews~\cite{8265649} & ['ind'] & ['News', 'Written'] & found & human-annotated & CC BY 4.0 \\
        & HSDNofaaulia~\cite{10.1145/3330482.3330491} & ['fil'] & ['Social', 'Written'] & found & human-annotated &  \\
        & IMDB~\cite{maas-EtAl:2011:ACL-HLT2011} & ['ind'] & ['Reviews', 'Written'] & found & human-annotated &  \\
        & Indonglish~\cite{Astuti2023} & ['ind'] & ['Social', 'Written'] & found & expert-annotated &  \\
        & JaDiIde~\cite{hidayatullah2020attention} & ['ind'] & ['Social', 'Written'] & found & derived &  \\
        & Karonese~\cite{Sitepu2024SentimentAI} & ['ind'] & ['Social', 'Web']  & found & derived &  \\
        & KhineMyanmarNews~\cite{khine2017automatic} & ['mya'] & ['News', 'Written'] & found & derived & GPL-3.0 \\
        & Krathu500 & ['tha'] & ['Social', 'Web', 'News', ...] & found & human-annotated &  \\
        & LazadaReview & ['fil'] & ['Reviews', 'Written'] & found & derived &  \\
        & LEMSentiment~\cite{DBLP:journals/corr/abs-2011-00677} & ['ind'] & ['Social', 'Review', 'Written'] & found & human-annotated & CC BY-SA 4.0 \\
        & LimeSoda~\cite{9678187} & ['tha'] & ['Healthcare', 'Written'] & found & human-annotated & CC BY 4.0 \\
        & MADLAD400~\cite{kudugunta2023madlad400} & ['tet'] & ['Web'] & found & derived & ODC-BY \\
        & MassiveIntent\textsuperscript{*}~\cite{fitzgerald2022massive} & ['ind', 'tha', 'vie', ...] & ['Spoken'] & found & human-annotated  & CC BY 4.0 \\
        & MassiveScenario\textsuperscript{*}~\cite{fitzgerald2022massive} & ['ind', 'tha', 'vie', ...] & ['Spoken'] & found & human-annotated  & CC BY 4.0 \\
        & Minang~\cite{koto2020minangkabau} & ['ind'] & ['Encyclopaedic', 'Written'] & found & derived & MIT \\
        & MultiLingualSentiment\textsuperscript{*}~\cite{mollanorozy-etal-2023-cross} & ['ind', 'tha', 'vie'] & ['Reviews', 'Written'] & found & derived &  \\
        & MurasuNews & ['tam'] & ['News', 'Written'] & found & derived & CC0 \\
        & News~\cite{khine2017automatic} & ['mya'] & ['News', 'Written'] & found & derived & GLP-3.0 \\
        & News & ['zsm'] & ['News', 'Written'] & found & derived &  \\
        & News & ['khm'] & ['Encyclopaedic', 'Web', 'News', ...] & found & derived &  \\
        & News & ['tam'] & ['News', 'Written'] & found & derived & CC BY-SA 4.0 \\
        & News~\cite{phatthiyaphaibun_2025_14967275} & ['lao'] & ['News', 'Written'] & found & derived &  \\
        & NewsDataset & ['ind'] & ['News', 'Written'] & found & derived &  \\
        & NusaX~\cite{winata-etal-2023-nusax} & ['ind'] & ['Social', 'Economics', 'Healthcare', ...] & found & expert-annotated & CC BY-SA 4.0 \\
        & PhoATIS~\cite{JointIDSF} & ['vie'] & ['Spoken'] & found & expert-annotated &  \\
        & PHElectionsSA & ['fil'] & ['Social'] & found & human-annotated &  \\
        & PHElectionsTD & ['fil'] & ['Social'] & found & human-annotated &  \\
        & Profanity~\cite{galinato-etal-2023-context} & ['fil'] & ['Social'] & found & human-annotated &  \\
        & ReviewShopping~\cite{phatthiyaphaibun-etal-2023-pythainlp} & ['tha'] & ['Reviews', 'Written'] & found & human-annotated & CC BY 3.0 \\
        & SIB200~\cite{adelani2023sib200} & ['ind', 'tha', 'vie', ...] & ['News', 'Written'] & found & expert-annotated & CC BY-SA 4.0 \\
        & SEATranslationeseResampled~\cite{lovenia2024seacrowd} & ['ind', 'tha', 'vie', ...] & ['News', 'Social', 'Culture', ...] & found & derived & Apache license 2.0 \\
        & SentEmoMobileApps~\cite{riccosan2023} & ['ind'] & ['Reviews', 'Written'] & found & human-annotated &  \\
        & SentimentAnalysis~\cite{ridife2019idsa} & ['ind'] & ['Social', 'Written'] & found & derived & CC BY-NC-ND 4.0 \\
        & ShopeeReviews\textsuperscript{*}~\cite{purwarianti2019improving} & ['fil'] & ['Social', 'Written'] & found & human-annotated & MLP-2.0 \\
        & SMSA & ['ind'] & ['Reviews', 'Written'] & found & derived & MIT \\
        & SpamidPair~\cite{Chrismanto2022} & ['ind'] & ['Social', 'Written'] & found & human-annotated & CC BY 4.0 \\
        & SpamReviews~\cite{10.1007/978-3-031-21743-2_48} & ['vie'] & ['Reviews', 'Written'] & found & human-annotated & CC BY-NC 4.0 \\
        & StudentFeedback~\cite{8573337} & ['vie'] & ['Reviews', 'Written'] & found & human-annotated & MIT \\
        & TCAS61~\cite{phatthiyaphaibun-etal-2023-pythainlp} & ['tha'] & ['Social', 'Written'] & found & human-annotated & CC BY 3.0 \\
        & The40ThaiChildrenStories~\cite{pasupa2016sentiment} & ['tha'] & ['Encyclopaedic', 'Written'] & found & human-annotated &  \\
        & ThuraMyanmarNews~\cite{aung2025news} & ['mya'] & ['News', 'Written'] & found & derived & MIT \\
        & TiktokHatespeech~\cite{hernandez2021bert} & ['fil'] & ['Social', 'Written'] & found & human-annotated & CC BY-SA 4.0 \\
        & Tweets~\cite{juan2022social} & ['zsm'] & ['Reviews', 'Written'] & found & derived &  \\
        & TyphoonYolandaTweets & ['fil'] & ['Social', 'Written'] & found & human-annotated & CC BY 4.0 \\
        & UITViCTSD~\cite{nguyen2021victsd} & ['vie'] & ['Social', 'Written'] & found & human-annotated &  \\
        & UITViHSD~\cite{10.1007/978-3-030-79457-6_35} & ['vie'] & ['Social', 'Written'] & found & human-annotated &  \\
        & UITViSFD~\cite{10.1007/978-3-030-82147-0_53} & ['vie'] & ['Social', 'Written'] & found & human-annotated &  \\
        & UITVION~\cite{fujita2021empirical} & ['vie'] & ['Social', 'Written'] & found & human-annotated &  \\
        & UITVSMEC~\cite{ho2020emotion} & ['vie'] & ['Social', 'Written'] & found & human-annotated &  \\
        & VaccinesTweets & ['ind'] & ['Social', 'Written'] & found & human-annotated &  \\
        & ViOCD~\cite{nguyen2021vietnamese} & ['vie'] & ['Reviews', 'Written'] & found & human-annotated &  \\
        & VLSP2016Sentiment~\cite{nguyen2018vlsp} & ['vie'] & ['Reviews', 'Written'] & found & human-annotated &  \\
        & WisesightSentiment~\cite{bact_2019_3457447} & ['tha'] & ['Social', 'News', 'Written'] & found & expert-annotated & CC0-1.0 \\
        & WongnaiReviews & ['tha'] & ['Reviews', 'Written'] & found & derived & LGPL-3.0 \\
        \midrule
        Multi-label Classification & BurmesePrachathai67k~\cite{phatthiyaphaibun-etal-2023-pythainlp} & ['mya'] & ['News', 'Web', 'Written'] & created & human-annotated & Apache license 2.0 \\
        & CASA~\cite{ilmania2018aspect} & ['ind'] & ['Reviews', 'Written'] & found & human-annotated & MIT \\
        & Dengue~\cite{8459963} & ['fil'] & ['Social', 'Written'] & found & derived & GLP-3.0 \\
        & GKLMIPNews~\cite{jiang2021khmer} & ['khm'] & ['News', 'Written'] & found & derived &  \\
        & HateSpeech~\cite{ibrohim-budi-2019-multi} & ['ind'] & ['Social', 'Written'] & found & human-annotated & CC BY-SA 4.0 \\
        & HoASA~\cite{azhar2019multi} & ['ind'] & ['Reviews', 'Written'] & found & human-annotated & MIT \\
        & Netifier~\cite{Izzan_Netifier_2025} & ['ind'] & ['Social', 'Written'] & found & human-annotated & CC BY-SA 4.0 \\
        & Prachathai67k~\cite{phatthiyaphaibun-etal-2023-pythainlp} & ['tha'] & ['News', 'Web', 'Written'] & found & derived & Apache license 2.0 \\
        & TrueVoiceIntent & ['tha'] & ['Conversation'] & found & derived &  \\
        & VLSP2018SAHotel~\cite{9865479} & ['vie'] & ['Reviews', 'Written'] & found & human-annotated &  \\
        & VLSP2018SARestaurant~\cite{9865479} & ['vie'] & ['Reviews', 'Written'] & found & human-annotated &  \\
        \midrule
        Pair Classification & BurmeseXNLI~\cite{conneau-etal-2018-xnli} & ['mya'] & ['Non-fiction', 'Fiction', 'Government'] & created & human-annotated & CC BY-NC 4.0 \\
        & IDKMRCNLI & ['ind'] & ['Encyclopaedic', 'News', 'Written'] & found &  &  \\
        & IndicXNLI\textsuperscript{*}~\cite{aggarwal_gupta_kunch_22} & ['tam'] & ['Non-fiction', 'Fiction', 'Government'] & found & expert-annotated & CC BY-NC 4.0 \\
        & IndoNLI\textsuperscript{*}~\cite{mahendra-etal-2021-indonli} & ['ind'] & ['Encyclopaedic', 'Web', 'News', ...] & found & expert-annotated & CC BY-SA 4.0 \\
        & MultilingualNLI26lang2mil7~\cite{laurer_less_2022} & ['ind', 'vie'] & ['Non-fiction', 'Fiction', 'Government'] & found & machine-translated and reviewed &  \\
        & MyXNLI~\cite{aung2024myanmarxnli} & ['mya'] & ['Non-fiction', 'Fiction', 'Government'] & found & human-annotated & CC BY-NC 4.0 \\
        & NewsPHNLI~\cite{cruz2020investigating} & ['fil'] & ['News', 'Written'] & found & human-annotated & GPL-3.0 \\
        & PAWS & ['fil'] & ['Web'] & found & human-annotated &  \\
        & SQuADNLI & ['ind'] & ['Encyclopaedic', 'News', 'Written'] & found &  &  \\
        & TyDIQANLI & ['ind'] & ['Encyclopaedic', 'News', 'Written'] & found &  &  \\
        & WReTE~\cite{setya2018semi} & ['tha'] & ['Encyclopaedic', 'Web', 'News'] & found & expert-annotated & MIT \\
        & XNLI\textsuperscript{*}~\cite{conneau-etal-2018-xnli} & ['tha', 'vie'] & ['Non-fiction', 'Fiction', 'Government'] & found & expert-annotated & CC BY-NC 4.0 \\
        & XNLITranslated~\cite{conneau-etal-2018-xnli} & ['khm', 'zsm', 'lao'] & ['Non-fiction', 'Fiction', 'Government'] & machine-translated and verified & machine-translated and reviewed & CC BY-NC 4.0 \\
        \bottomrule
    \end{tabular}
    }
    \caption{The datasets included in SEA-BED (part 1).}
    \label{tab:datasets_in_sea-mteb_part1}
\end{table*}

\begin{table*}[h!]
\hspace{-3mm}
\centering
\setlength\doublerulesep{4pt}
\scalebox{0.4}{
    \renewcommand{\arraystretch}{1.2}
    \begin{tabular}{lllllll}
        \toprule
        \textbf{Type} & \textbf{Name} & \textbf{Languages} & \textbf{Domains} & \textbf{Sample creation} & \textbf{Annotations creators} & \textbf{License}
        \\
        \bottomrule
         STS & Biosses~\cite{souganciouglu2017biosses} & ['tha', 'mya'] & ['Medical'] & created & human-annotated & GPL-3.0 \\
        & BiossesCrosslingual~\cite{souganciouglu2017biosses} & ['tha', 'mya'] & ['Medical'] & created & human-annotated & GPL-3.0 \\
        & IndicCrosslingual\textsuperscript{*}~\cite{DBLP:journals/tacl/RameshDBJASSDJK22} & ['tam'] & [News, Non-fiction, Web, ...] & found & expert-annotated & CC0-1.0 \\
        & SemRel2024\textsuperscript{*}~\cite{ousidhoum2024semrel2024} & ['ind'] & ['Spoken',  'Written'] & found & human-annotated &  \\
        & STS17~\cite{cer-etal-2017-semeval} & ['tha', 'mya'] & ['News', 'Web', 'Written'] & created & human-annotated &  \\
        & STS17Crosslingual~\cite{cer-etal-2017-semeval} & ['tha', 'mya'] & ['News', 'Web', 'Written'] & created & human-annotated &  \\
        & STS22~\cite{chen-etal-2022-semeval} & ['tha', 'mya'] & ['News', 'Written'] & created & human-annotated &  \\
        & STS22Crosslingual~\cite{chen-etal-2022-semeval} & ['tha', 'mya'] & ['News', 'Written'] & created & human-annotated &  \\
        & STS24~\cite{ousidhoum-etal-2024-semeval} & ['tha', 'mya'] & ['Spoken',  'Written'] & created & human-annotated &  \\
        & STS24Crosslingual~\cite{ousidhoum-etal-2024-semeval} &['tha', 'mya'] & ['Spoken',  'Written'] & created & human-annotated &  \\
        & STSBenchmark~\cite{cer-etal-2017-semeval} & ['ind', 'tha', 'vie', ...] & ['News', 'Web',  'Written'] & machine-translated and verified & machine-translated and reviewed & CC BY-SA 4.0 \\
        \midrule
        Clustering & EMoTES3K~\cite{catapang-visperas-2023-emotion} & ['fil'] & ['Morality', 'Written'] & found & human-annotated & Apache license 2.0 \\
        & MurasuNews & ['tam'] & ['News', 'Written'] & found & derived & CC0 \\
        & News~\cite{phatthiyaphaibun_2025_14967275} & ['lao'] & ['News', 'Written'] & found & derived &  \\
        & News~\cite{9645441} & ['khm'] & ['News', 'Written'] & found & derived &  \\
        & News~\cite{andreaschandra2020} & ['ind'] & ['News', 'Written'] & found & derived &  \\
        & News & ['tam'] & ['News', 'Written'] & found & derived & CC BY-SA 4.0 \\
        & News~\cite{khine2017automatic} & ['mya'] & ['News', 'Written'] & found & derived &  \\
        & SIB200~\cite{adelani2023sib200} & ['ind', 'tha', 'vie', ...] & ['News', 'Written'] & found & expert-annotated & CC BY-SA 4.0 \\
        & UITVION~\cite{fujita2021empirical} & ['vie'] & ['Social', 'Written'] & found & human-annotated &  \\
        & ViOCD~\cite{nguyen2021vietnamese} & ['vie'] & ['Reviews', 'Written'] & found & human-annotated &  \\
        \midrule
        BitextMining & ALT~\cite{riza2019asian} & ['ind', 'tha', ...] & ['News', 'Written'] & found & expert-annotated & CC BY 4.0 \\
         & BibleNLP\textsuperscript{*}~\cite{akerman2023ebible} & ['ind', 'tha', 'vie', ...] & ['Religious', 'Written'] & found & expert-annotated & CC BY 4.0 \\
         & Flores\textsuperscript{*}~\cite{goyal2022flores} & ['ind', 'tha', 'vie', ...] & ['Non-fiction', 'Encyclopaedic', 'Written'] & found & human-annotated &  \\
         & Embassy~\cite{wannaphong2020pythainlp} & ['tha', 'lao'] & ['Government', 'News'] & found & human-annotated & CC0-1.0 \\
         & IN22Conv\textsuperscript{*}~\cite{gala2023indictrans}& ['tam'] & ['Social', 'Spoken', 'Fiction', ...] & found & expert-annotated & CC BY 4.0 \\
         & IN22Gen\textsuperscript{*}~\cite{gala2023indictrans} & ['tam'] & ['Web', 'Legal', 'Government', ...] & found & expert-annotated & CC BY 4.0 \\
         & IndoGeneral~\cite{guntara-etal-2020-benchmarking} & ['ind'] & ['General', 'Writen'] & found & derived & CC BY-SA 4.0 \\
         & IndoIdentic~\cite{gala2023indictrans} & ['ind'] & ['News', 'Spoken', 'Web', ...] & found & derived &  \\
         & IndoNLG~\cite{cahyawijaya-etal-2021-indonlg} & ['ind'] & ['religion'] & found & derived &  \\
         & IndoNews~\cite{guntara-etal-2020-benchmarking} & ['ind'] & ['News', 'Written'] & found & derived & CC BY-SA 4.0 \\
         & IndoReligious~\cite{guntara-etal-2020-benchmarking} & ['ind'] & ['Religion', 'Writen'] & found & derived & CC BY-SA 4.0 \\
         & Liputan6~\cite{koto2020liputan6} & ['ind'] & ['News', 'Written'] & found & human-annotated & CC BY-SA 4.0 \\
         & MADLAD400~\cite{kudugunta2023madlad400} & ['tet'] & ['Web'] & found & derived & ODC-BY \\
         & NTREX\textsuperscript{*}~\cite{federmann-etal-2022-ntrex} & ['ind', 'tha', 'vie', ...] & ['News', 'Written'] & found & expert-annotated & CC BY-SA 4.0 \\
         & NusaxMiners\textsuperscript{*}~\cite{winata-etal-2023-nusax} & ['ind'] & ['Reviews', 'Written'] & found & human-annotated & CC BY-SA 4.0 \\
         & QED~\cite{lamm2020qed} & ['ind', 'tha', 'vie', ...] & ['Education', 'Social', 'Spoken', ...] & found & human-annotated & CC BY-SA \\
         & SCBMTEnTh2020~\cite{DBLP:journals/lre/LowphansirikulP22} & ['tha'] & ['conversation', 'Web', 'Government', ...] & found & human-annotated & CC BY-SA 4.0 \\
         & SoftwareDocumentation~\cite{buschbeck2020parallelevaluationdataset} & ['ind', 'tha', 'vie', ...] & ['Web', 'Product'] & found & expert-annotated & CC BY-NC 4.0 \\
         & TALPCo~\cite{published_papers/22434604, published_papers/22434603} & ['ind', 'tha', 'vie', ...] & ['Conversation', 'spoken'] & found & human-annotated & CC BY-4.0 \\
         & Tatoeba\textsuperscript{*}~\cite{tiedemann-2020-tatoeba} & ['ind', 'tha', 'vie', ...] & ['Written'] & found & human-annotated & CC BY-2.0 \\
         & TED2020~\cite{reimers-2020-multilingual-sentence-bert} & ['ind', 'tha', 'vie', ...] & ['Education', 'Social', 'Spoken', ...] & found & human-annotated & CC BY-NC-ND 4.0 \\
         & ThaiGov & ['tha'] & ['Government', 'News'] & found & human-annotated & PDDL \\
         & USEmbassy~\cite{phatthiyaphaibun-etal-2023-pythainlp} & ['tha'] & ['News'] & found & derived & CC0-1.0 \\
         & VSoLSCSum~\cite{nguyen-etal-2016-vsolscsum} & ['vie'] & ['Social', 'Written'] & found & human-annotated & CC BY-4.0 \\
         & XLSum~\cite{hasan2021xl} & ['ind', 'tha', 'vie', ...] & ['News', 'Written'] & found & human-annotated & CC BY-NC-SA 4.0 \\
        \midrule
        Retrieval & ACIQuAD~\cite{afa5bf8149d6406786539c1ea827087d} & ['ind'] & ['Encyclopaedic', 'Written'] & found & expert-annotated & CC-BY 4.0 \\
        & Agricutlure1K~\cite{Myanmar-Agriculture-1K} & ['mya'] & ['Encyclopaedic', 'Written'] & found & expert-annotated & CC BY-SA 4.0 \\
        & AskCovidDrBot~\cite{aung2025askcoviddrbot} & ['mya'] & ['Encyclopaedic', 'Written'] & found & human-annotated & MIT \\
        & ChatGPTOpenQA & ['zsm'] & ['Encyclopaedic', 'Written'] & found & LM-generated & CC BY-NC-SA 2.0 \\
        & ContextSearch~\cite{nguyen2025advancingvietnameseinformationretrieval} & ['tha'] & ['STEM', 'Humanities', 'Social Sciences', ...] & found & human-annotated & MIT \\
        & IAppWiki~\cite{kobkrit_viriyayudhakorn_2021_4539916} & ['tha'] & ['Encyclopaedic', 'Web', 'News'] & found & expert-annotated & MIT \\
        & IDKMRC~\cite{putri-oh-2022-idk} & ['tnd'] & ['Encyclopaedic', 'Written'] & found & human-annotated & CC BY-SA 4.0 \\
        & IndicQA\textsuperscript{*}~\cite{doddapaneni2022towards} & ['tam'] & ['Web', 'Written'] & machine-translated and verified & human-annotated & CC BY 4.0 \\
        & IndoNLG~\cite{cahyawijaya-etal-2021-indonlg} & ['ind'] & ['Religion', 'Writen'] & found & human-annotated & CC BY-SA 4.0 \\
        & IndoQA~\cite{IndoQA} & ['ind'] & ['Web'] & found & expert-annotated & CC BY-ND 4.0 \\
        & MLDR~\cite{bge-m3} & ['tha'] & ['Encyclopaedic', 'Written'] & found & LM-generated & MIT \\
        & MLQA\textsuperscript{*}~\cite{lewis2019mlqa} & ['vie'] & ['Encyclopaedic', 'Written'] & found & human-annotated & CC BY-SA 3.0 \\
        & MIRACL\textsuperscript{*}~\cite{10.1162/tacl_a_00595} & ['ind', 'tha'] & ['Encyclopaedic', 'Written'] & found & expert-annotated & Apache license 2.0 \\
        & Microbiology1K~\cite{si_thu_2024_burmese} & ['mya'] & ['Encyclopaedic', 'Written'] & found & human-annotated & CC BY-SA 4.0 \\
        & QASiNa~\cite{10390123} & ['ind'] & ['Religion', 'Writen'] & found & human-annotated & MIT \\
        & ThaiWikiQA~\cite{trakultaweekoon2019first} & ['tha'] & ['Encyclopaedic', 'Written'] & found & human-annotated & CC BY-NC-SA 3.0 \\
        & TyDiQA~\cite{tydiqa} & ['ind', 'tha'] & ['Encyclopaedic', 'Written'] & found & human-annotated & Apache license 2.0 \\
        & ViQuAD2\_0~\cite{Nguyen_2022} & ['vie'] & ['Encyclopaedic', 'Written'] & found & expert-annotated & MIT \\
        & WangchanXLegalThaiCCLRAG~\cite{akarajaradwong2025nitibenchcomprehensivestudiesllm} & ['tha'] & ['Legal', 'Written'] & found & human-annotated & MIT \\
        & XQuAD~\cite{Artetxe:etal:2019} & ['tha', 'vie'] & ['Web', 'Written'] & found & human-annotated & CC BY-SA 4.0 \\
        \midrule
        Instruction Retrieval & AlpacaInstruct & ['ind'] & None & found & LM-generated & Apache license 2.0 \\
        & Vietnamese52KAlpaca~\cite{vietnameseLLM} & ['vie'] & None & found & LM-generated &  \\
        & WangchanThaiInstruct & ['tha'] & ['Medical', 'Finance', 'Legal', ...] & found & human-annotated & CC BY-SA 4.0 \\
        & WangchanXSyntheticInstructThai120k~\cite{pengpun-etal-2024-seed} & ['tha'] & ['Encyclopaedic', 'Written'] & found & LM-generated & MIT \\
        \midrule
        Reranking & MIRACL~\cite{10.1162/tacl_a_00595} & ['ind', 'tha'] & ['Encyclopaedic', 'Written'] & found & expert-annotated & Apache license 2.0 \\
        \bottomrule
    \end{tabular}
    }
    \caption{The datasets included in SEA-BED (part 2).}
    \label{tab:datasets_in_sea-mteb_part2}
\end{table*}